\pdfoutput=1
\documentclass{article}
\usepackage{iclr2027_conference}
\usepackage[T1]{fontenc}
\renewcommand{\sfdefault}{phv}
\font\iclrtenhv=cmr8 at 8pt

\usepackage{amsmath,amsfonts,bm}

\def\eqref#1{equation~\ref{#1}}

\def\1{\bm{1}}

\DeclareMathAlphabet{\mathsfit}{\encodingdefault}{\sfdefault}{m}{sl}
\SetMathAlphabet{\mathsfit}{bold}{\encodingdefault}{\sfdefault}{bx}{n}

\usepackage{booktabs}
\usepackage{array}
\usepackage{graphicx}
\usepackage{float}
\usepackage{tikz}
\usetikzlibrary{arrows.meta,positioning,fit,calc,shapes.geometric}
\usepackage{microtype}
\usepackage{enumitem}
\usepackage{algorithm}
\usepackage{algpseudocode}
\usepackage{xcolor}
\usepackage[hidelinks]{hyperref}
\usepackage{url}

\title{\fontsize{14.4}{17}\selectfont IntentTwin: Deconfounding Candidate Support\\and the Oracle-to-Deployment Gap\\in Offline Web Behavior Prediction}

\author{Miaobo Hu$^{1,2}$, Shuhao Hu$^{2}$, Xiaobo Guo$^{2}$, Xin Wang$^{2}$,\\
Bokun Wang$^{2}$, Daren Zha$^{2}$, Jun Xiao$^{1,*}$\\[4pt]
{\normalfont $^{1}$School of Artificial Intelligence, University of Chinese Academy of Sciences, Beijing, China}\\
{\normalfont $^{2}$Institute of Information Engineering, Chinese Academy of Sciences, Beijing, China}\\
{\normalfont $^{*}$Corresponding author: \texttt{xiaojun@ucas.ac.cn}}}

\newif\ifcontextsweepavailable
\contextsweepavailablefalse
\IfFileExists{tables/sft_context_length_sweep_s13.tex}{%
  \IfFileExists{figures/context_length_sweep.pdf}{%
    \contextsweepavailabletrue
  }{}
}{}

\iclrfinalcopy
\begin{document}
\raggedbottom
\maketitle
\lhead{Preprint}

\begin{abstract}
Offline web-behavior benchmarks frequently insert the demonstrated action into
a finite candidate set and then measure reranking. We show that this protocol
can create a candidate-support confound: when the target is outside training
support while all distractors are sampled from that support, support membership
identifies the gold action without modeling behavior. We formalize this effect
with support-identifiability rate (SIR), construct support-balanced candidate
sets, and evaluate a target-free dual-path proposer that combines support
retrieval with open-vocabulary DOM/LM proposals.

On filtered OPeRA (4,282/469/992 train/dev/test steps), the original protocol
has SIR $1.0000$ on target-OOV rows and a support-only rule reaches $1.0000$
OOV Exact@1. Support balancing reduces SIR to $0.0000$; the strongest
conditional reranker obtains $0.7214$ overall Exact@1, $0.7988$ on IV rows,
and $0.5668$ on OOV rows. The target-free dual-path system reaches Proposal
Recall@16 $0.6915$ and end-to-end Exact@1 $0.5212$, compared with oracle
conditional Exact@1 $0.7450$. On Mind2Web, the same audit yields SIR $0.2380$
and an oracle-to-proposal gap of $0.2125$. A 200-row semantic audit labels
$0.7150$ of exact-key OOV targets as semantically novel, $0.2450$ as
locator/string variation, and $0.0400$ as ambiguous (Cohen's $\kappa=0.8173$).
These measurements show why support composition must be controlled alongside
candidate accuracy when offline results are used to reason about deployment.
\end{abstract}

\section{Introduction}
Web interaction traces contain a prefix of observations and actions, but an
offline benchmark often supplies the gold action as an item in the candidate
set. Consider a target such as \emph{click ``Blue 256GB''}. In an oracle
evaluation it is guaranteed to appear among 15 distractors; in deployment, a
system must first discover and name that action from the current page and
history. Ranking correctness is therefore conditional on candidate
availability.

A second issue appears when candidate availability is correlated with the
evaluation label. In the original target-anchored construction, the logged
action is inserted into the candidate set while every negative is sampled from
the frozen training vocabulary. For a target-OOV row, the gold action is thus
the unique out-of-support candidate. A support-only rule can identify the
target from candidate membership alone; we call this
\emph{candidate-support confounding}. We therefore separate behavioral
discrimination from the information revealed by the support composition of the
candidate set.

We study this distinction with a target-free context $c_t$ and a structured
action key $a_t=(\mathrm{type},\mathrm{target},\mathrm{value})$. The protocol
freezes the candidate support using training targets, keeps sessions
disjoint, and stores an ordered row ledger containing candidates,
probabilities, ranks, NLLs, and provenance. A test target whose exact key is
absent from the training support is marked target-OOV; it is never added to the
training vocabulary or the model input.

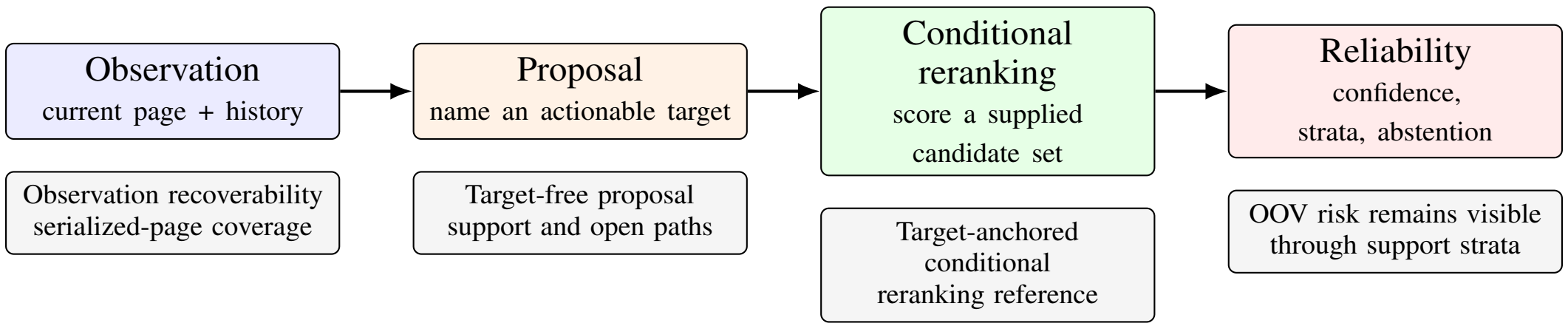
\begin{figure*}[t]
  \centering
  \begin{tikzpicture}[
    box/.style={draw, rounded corners=2pt, align=center, minimum height=0.62cm,
      text width=2.75cm, font=\small},
    metric/.style={draw, rounded corners=2pt, fill=gray!8, align=center,
      minimum height=0.58cm, text width=2.75cm, font=\scriptsize},
    arrow/.style={-{Latex[length=2mm]}, thick}
  ]
    \node[box, fill=blue!8] (obs) {Observation\\\scriptsize current page + history};
    \node[box, fill=orange!10, right=0.65cm of obs] (prop) {Proposal\\\scriptsize name an actionable target};
    \node[box, fill=green!10, right=0.65cm of prop] (rank) {Conditional reranking\\\scriptsize score a supplied candidate set};
    \node[box, fill=red!8, right=0.65cm of rank] (rel) {Reliability\\\scriptsize confidence, strata, abstention};
    \draw[arrow] (obs) -- (prop);
    \draw[arrow] (prop) -- (rank);
    \draw[arrow] (rank) -- (rel);
    \node[metric, below=0.28cm of obs] {Observation recoverability\\serialized-page coverage};
    \node[metric, below=0.28cm of prop] {Target-free proposal\\support and open paths};
    \node[metric, below=0.28cm of rank] {Target-anchored conditional\\reranking reference};
    \node[metric, below=0.28cm of rel] {OOV risk remains visible\\through support strata};
  \end{tikzpicture}
  \caption{The evaluation contract separates observation, target-free proposal,
  conditional reranking, and reliability. Each regime has its own candidate
  set and denominator; the end-to-end result is computed only from the
  executable proposal--reranking pipeline.}
  \label{fig:evaluation-contract}
\end{figure*}

We organize the study around four questions:
\begin{enumerate}[leftmargin=1.4em]
  \item \textbf{Support confounding:} When does candidate membership reveal
        the gold action independently of the behavioral context?
  \item \textbf{Deconfounded reranking:} How do rankers behave when IV and OOV
        candidates are support-balanced?
  \item \textbf{Oracle-to-deployment gap:} What changes when the oracle anchor
        is replaced by a target-free proposer?
  \item \textbf{Support-aware proposal:} Can context-gated retrieval and
        open-vocabulary proposal improve coverage without candidate-level gates?
\end{enumerate}

Our contributions are:
\begin{itemize}[leftmargin=1.4em]
  \item a formal candidate-support confound and the support-identifiability
        rate (SIR) diagnostic;
  \item a support-balanced candidate protocol with exact action-key strata,
        pseudo-OOV calibration, and complete candidate-level ledgers; and
  \item a target-free dual-path proposal evaluation that reports proposal
        recall, covered reranking, and end-to-end exact action selection on
        OPeRA and Mind2Web.
\end{itemize}

\section{Related Work}
\paragraph{Offline decision modeling.}
Maximum-entropy inverse reinforcement learning (MaxEnt IRL) models a
demonstration as a distribution over trajectories rather than as a deterministic
label \citep{ziebart2008maximum}. Our scorer uses a finite conditional
candidate-level softmax objective rather than a trajectory partition function.
We therefore refer to this family as factorized candidate scoring below; some
implementation artifacts retain the historical ``IRL'' label. The objective is
an IRL-inspired finite-choice energy surrogate: it does not solve a
trajectory-level inverse problem, occupancy-matching problem, or reward-
identification problem. Offline RL methods address distribution shift
by constraining learned values to the support of logged data, with implicit
value learning providing a useful reference point for our candidate-ranking
setting \citep{levine2020offline,kumar2020cql,kostrikov2022iql}. Dataset and
protocol design are central to offline evaluation, as emphasized by D4RL
\citep{fu2021d4rl}. Diffusion-based trajectory models offer a
complementary way to represent multi-step behavior, but their offline gains
depend strongly on the quality of the latent and action interfaces
\citep{janner2022diffuser}. We borrow the probabilistic decision view while
making the candidate support and preference factors explicit.

\paragraph{Web and tool-using agents.}
ReAct interleaves reasoning and acting \citep{yao2023react}, while Toolformer
and Reflexion study tool use and verbal feedback
\citep{schick2023toolformer,shinn2023reflexion}. WebArena and Mind2Web provide
realistic browser tasks and traces \citep{zhou2024webarena,deng2023mind2web};
AgentBench broadens evaluation across agent environments
\citep{liu2024agentbench}. These benchmarks motivate our focus on action
semantics and session-level context. Mind2Web in particular uses candidate
generation and top-$k$ recall, whereas our narrower contribution is the
train-only candidate contract, row-level likelihood ledger, and explicit
target-OOV analysis for a closed-choice factorized scorer.

\paragraph{Preference learning and explanations.}
Direct preference optimization and human-feedback fine-tuning
\citep{rafailov2023dpo,ouyang2022training} are useful future baselines for our
language-model arm, whereas the present mainline isolates a symbolic candidate
scorer. Self-Refine and Voyager show how feedback and self-generated memory can
improve agent behavior \citep{madaan2023selfrefine,wang2023voyager}; our focus
is the preceding measurement contract that makes such improvements auditable.
Finally, the faithfulness literature warns against treating textual rationales
as explanations without independent validation \citep{turpin2023unfaithful}.

\section{Method}
\label{sec:method}
\subsection{Evaluation contract and candidate support}
Let a session be a sequence of observed steps
$\tau=(o_1,a_1,\ldots,o_T,a_T)$, where each action $a_t$ is serialized as an
action type and a structured target. For a prediction row at time $t$, the
context $c_t$ contains only the prefix observations and actions, mode-selected
identity fields (user ID and/or persona when enabled), and metadata available
before $a_t$. The target $a_t$ is excluded from $c_t$. We score a candidate set
$\mathcal{A}_t$ of size 16 and report
\[
  p_\theta(a\mid c_t)=
  \frac{\exp(R_\theta(c_t,a)/\beta)}
  {\sum_{a'\in\mathcal{A}_t}\exp(R_\theta(c_t,a')/\beta)},
  \qquad \beta=0.5.
\]
The central distinction is between the action support available to a scorer and
the action that was logged. Let
\[
  V_{\mathrm{train}}=\{(\mathrm{type},\mathrm{target},\mathrm{value})
  \text{ keys observed in }D_{\mathrm{train}}\}.
\]
An evaluation row is IV when its exact target key is in
$V_{\mathrm{train}}$, and target-OOV otherwise. For example, ranking a logged
``Blue 256GB'' click among supplied alternatives tests conditional reranking;
deployment must first propose and name that target. Ranking correctness is
therefore conditional on candidate availability.
The negative-action vocabulary is frozen from training action keys. For each
row, the logged target is inserted once as an evaluation anchor and 15
deterministic negatives are drawn from that vocabulary; the configured ratio
selects up to eight hard negatives (sharing the action type or target when
possible) and seven random non-duplicates, with a deterministic small-pool
fallback. Thus a test target outside the
training vocabulary is scoreable but marked target-OOV: it does not update the
global vocabulary or supply input context. This target-anchored design has
100\% candidate-anchor coverage by construction and evaluates conditional
reranking under a controlled choice set; it is not open-vocabulary action
generation. The evaluator inserts the logged target only after the model has
formed the target-free context, so the reported Top-$k$ values measure
conditional reranking given an oracle candidate anchor rather than the
probability of proposing that action in deployment.

The builder deduplicates the train action vocabulary by the
$(\mathrm{type},\mathrm{target},\mathrm{value})$ key before sampling. With 16
slots and a hard-negative ratio of 0.5, it requests up to eight same-type or
same-target negatives and fills the remainder with random train-vocabulary
actions. A SHA-256 digest of
\texttt{session\_id:step\_id:seed} initializes the per-row RNG, and a small
pool falls back to unused hard negatives without duplicating the anchor. This
makes the candidate list byte-reproducible while exposing a controlled source
of difficulty: the held-out rows contain 777 click rows with 15 same-type
negatives, 65 with 14, 3 with 13, 107 input rows with two, and 40 terminate
rows with none. The resulting frozen train vocabulary contains 1,280 unique
action keys; its digest is retained in the split statistics and every run
receipt.

\paragraph{Support identifiability and balanced candidates.}
For a candidate set $\mathcal{A}_t$, let $I(a)=\mathbb{1}[a\in V_{\mathrm{train}}]$.
The support-identifiability rate (SIR) is the fraction of rows for which the
gold candidate is the unique candidate with its support status. The support-only
baseline observes $I(a)$ and selects that unique candidate, breaking ties by
the frozen frequency order. To remove this cue, the support-balanced protocol
fixes $q$ out-of-support candidates and $16-q$ in-support candidates in every
set, with $q\in\{1,2,4,8\}$; $q=4$ is the primary setting. Balanced candidates
preserve the action-type hard-negative ratio and use the same deterministic row
seed.

\subsection{Three evaluation regimes}
The same target-free context is evaluated under three deliberately different
information contracts. (1) \textbf{Oracle candidate reranking} inserts the
logged target into a fixed 16-way set and measures Top-1, NLL, MRR, exact-action
accuracy, action-type accuracy, and Macro-F1. (2) \textbf{Closed-vocabulary
proposal} removes the target anchor and original candidate list, then ranks
the complete frozen $V_{\mathrm{train}}$; Recall@$k$ and Exact@1 measure the
proposal boundary, and target-OOV recall is structurally zero. (3)
\textbf{Observation recoverability} extracts named actions from the current
serialized observation without reading the target or source candidate list.
Its coverage is an information ceiling, not a learned generator. We never
multiply metrics across regimes: each number is conditional on its own
candidate and observation assumptions.

\begin{figure*}[t]
  \centering
  \includegraphics[width=0.98\textwidth]{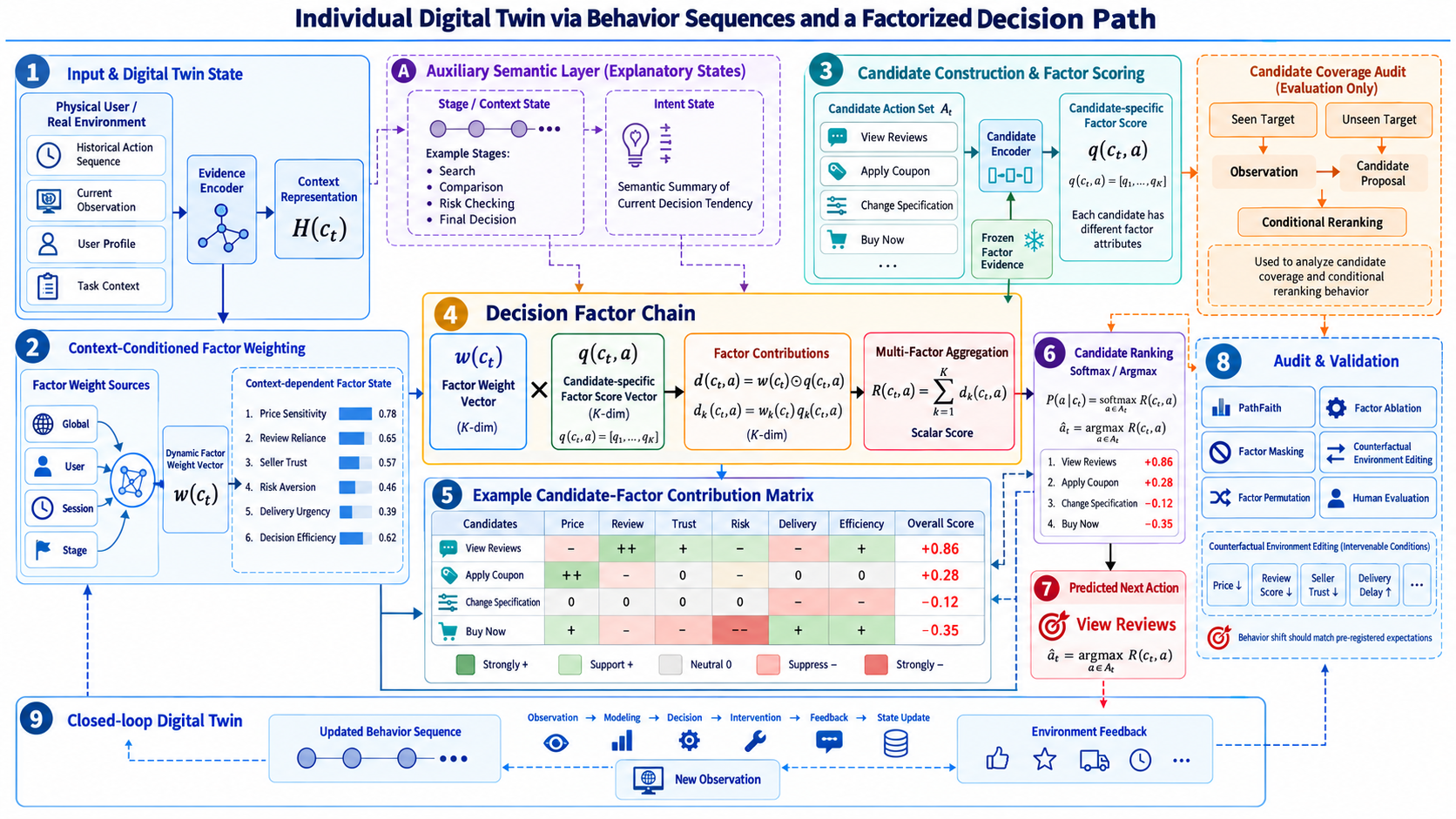}
  \caption{Algorithm overview used for the factorized decision path. The
  context encoder produces $H(c_t)$ and factor weights; candidate-specific
  evidence is combined into $R(c_t,a)$, ranked, and passed to the audit. The
  support audit separates seen and unseen targets before conditional reranking,
  while the proposal path supplies candidates without the logged target.}
  \label{fig:algorithm}
\end{figure*}

\subsection{Data protocol}
We use the filtered OPeRA protocol \citep{wang2025opera}: 394/43/90
session-disjoint train/dev/test sessions and 4,282/469/992 steps. The test
split contains 661 IV and 331 target-OOV rows (845 click, 107 input, and 40
terminate). Users recur across splits, so the primary claim is session-level
generalization rather than unseen-user transfer. A deterministic user-disjoint
stress test (3,799/1,419/525 rows; 187 target-OOV test rows) is retained only
as a secondary diagnostic. All paper-facing claims use the session-disjoint
split unless stated otherwise.

\subsection{Information timing}
The target-free context flows through the model; the demonstrated action enters
only the training loss or the held-out evaluator after that context is formed.
Train-fitted vocabularies, candidate lists, and optional segmentation or
representation artifacts are frozen before dev/test. Every held-out row retains
its ordered candidates, probabilities, rank, NLL, and provenance hashes.

\subsection{Scorers and support-aware hybrid}
The protocol-matched Candidate-CE control maps target-free context and
canonical action features to one scalar logit per candidate. Factorized
conditions use the same 16-way softmax but parameterize the score as
$R(c,a)=w(c)^\top q(c,a)$ with a train-frozen 14-factor bank and lightweight
hash features. These coordinates are auditable model states, not identified or
human-validated preferences; implementation details and secondary probes are
reported in Appendix~\ref{app:additional}.

The observed IV/OOV trade-off motivates a gate that uses no held-out targets.
Combining a target-free prior $s_{\mathrm{prior}}(c,a)$ with a candidate-ranker
score gives
\[
 s_{\mathrm{hyb}}(c,a)=s_{\mathrm{prior}}(c,a)+
 \alpha g(a)s_{\mathrm{rank}}(c,a),\qquad
 g(a)=\mathbb{1}\{\mathrm{count}_{D_{\mathrm{train}}}(a)\geq m\}.
\]
For the reported diagnostic, $m=1$: an action key is in-support if it appears
at least once in the training split. The gate therefore uses only training
support counts and does not inspect the held-out target.
We withhold 20\% of training action keys as pseudo-OOV keys, fit the gate and
$\alpha$ without test targets, and score the held-out test split once. This is
a diagnostic of support-aware interpolation, not a claim of open-vocabulary
generation.

\paragraph{Finite-choice objective.}
The scorer optimizes finite-choice cross-entropy, optionally with smoothness,
hierarchy, and masked rationale penalties. This is an IRL-inspired energy
surrogate rather than trajectory-level MaxEnt IRL: it has no trajectory
partition function, occupancy matching, reward identification, or sequential
credit assignment. Every inference row retains its ordered candidates,
probabilities, rank, NLL, stratum, and run hashes.

\section{Experiments}
\subsection{Setup and evaluation protocol}
\paragraph{Reproducible setup.}
All conditions use the same session-disjoint filtered OPeRA split, frozen
16-way candidate lists, target-free prompts, and train-only support artifacts.
Factorized controls use three epochs, learning rate $2\times10^{-5}$, gradient
accumulation 16, temperature 0.5, and seeds $13,42,2026$. The QLoRA reference
uses Qwen3-4B-Instruct-2507 with 4-bit adapters on $q_{\mathrm{proj}}$ and
$v_{\mathrm{proj}}$, rank 16, three epochs over all 4,282 training rows, and
2,048-token training contexts. All six full-budget adapters have complete
992-row ledgers under both the matched 384-token held-out evaluator and the
2,048-token full-context evaluator. The matched evaluation and detailed
receipts are in Appendix~\ref{app:additional}; the full-context persona
follow-up is a separate single-condition aggregate.
No test row is used for model selection, and completion NLL is not compared
with candidate-softmax NLL.

\paragraph{Metrics.}
We report candidate Top-1, target NLL, MRR, exact-action accuracy, action-type
accuracy, Macro-F1, and IV/OOV strata. Candidate Top-1 equals exact-action
accuracy because the target is inserted as an anchor. Where multiple seeds are
available, means $\pm$ sample SD are computed across training seeds; single-seed
rows are shown without dispersion. Row-level ledgers and clean run metadata bind
every reported aggregate.

\subsection{Support confounding and balanced reranking}
The original target-anchored protocol makes support membership observable at the
candidate level. Every target-OOV row contains one OOV gold action and
15 in-support negatives, so the support-only baseline obtains OOV Exact@1
$1.0000$ by construction. We report SIR together with behavioral scores to
separate this shortcut from context-sensitive ranking.

\begin{table}[!ht]
\centering
\scriptsize
\begin{tabular}{llccc}
\toprule
Protocol & Stratum & SIR & Support-only & QLoRA \\
\midrule
Original anchor & Overall & 0.3337 & 0.3760 & \textbf{0.7450} \\
Original anchor & IV      & 0.0000 & 0.0635 & \textbf{0.8200} \\
Original anchor & OOV     & \textbf{1.0000} & \textbf{1.0000} & 0.5952 \\
\midrule
Balanced ($q=4$) & Overall & 0.0000 & 0.0847 & \textbf{0.7214} \\
Balanced ($q=4$) & IV      & 0.0000 & 0.0000 & \textbf{0.7988} \\
Balanced ($q=4$) & OOV     & 0.0000 & 0.2538 & \textbf{0.5668} \\
\bottomrule
\end{tabular}
\caption{Candidate-support confounding on the complete 992-row test split.
SIR is the fraction of rows whose gold action is uniquely identified by support
status; Support-only uses no context. QLoRA is the matched 2,048-token
conditional reranker. Bold marks the best value within each stratum.}
\label{tab:support-confounding}
\end{table}

Support balancing reduces SIR from $1.0000$ to $0.0000$ on target-OOV rows and
reduces the support-only OOV score to $0.2538$. The QLoRA reranker retains
$0.5668$ OOV Exact@1 and $0.7214$ overall Exact@1, showing that the deconfounded
protocol still measures context-dependent discrimination.

\subsection{Support-balance sensitivity}
We vary the number $q$ of out-of-support candidates while fixing the candidate
budget at 16. The $q=4$ setting is primary; the sweep tests whether the
observed ranking changes are tied to one candidate composition.

\begin{table}[!ht]
\centering
\scriptsize
\begin{tabular}{ccccc}
\toprule
$q$ & SIR & Overall & IV & OOV \\
\midrule
1 & 0.3337 & \textbf{0.7429} & \textbf{0.8154} & \textbf{0.5980} \\
2 & 0.0000 & 0.7319 & 0.8079 & 0.5801 \\
4 & 0.0000 & 0.7214 & 0.7988 & 0.5668 \\
8 & 0.0000 & 0.6951 & 0.7791 & 0.5274 \\
\bottomrule
\end{tabular}
\caption{Sensitivity to the number of out-of-support candidates. SIR is lower
when support status is less informative; Exact@1 is reported overall and by
support stratum. Bold marks the best Exact@1 in the sweep.}
\label{tab:q-sweep}
\end{table}

Increasing $q$ removes the unique-OOV cue while preserving the fixed candidate
budget. Overall Exact@1 changes from $0.7429$ at $q=1$ to $0.7214$ at $q=4$
and $0.6951$ at $q=8$; the corresponding OOV values are $0.5980$, $0.5668$,
and $0.5274$.

\subsection{Target-free proposal and end-to-end action prediction}
We remove the demonstrated action and the source candidate list. The proposal
stage compares train-frequency retrieval, support retrieval, DOM-copy proposal,
constrained QLoRA generation, a fixed union, and the context-gated dual path.
Conditional Exact@1 is computed only when the proposal contains the exact
target; E2E Exact@1 uses all 992 rows.

\begin{table*}[t]
\centering
\scriptsize
\resizebox{\textwidth}{!}{\begin{tabular}{lrrrrrrr}
\toprule
Method & R@1 & R@4 & R@16 & R@128 & Cond. & E2E & OOV E2E \\
\midrule
Train frequency & 0.1079 & 0.3105 & 0.5151 & 0.6089 & 0.7045 & 0.3629 & 0.0000 \\
Candidate-CE support retrieval & 0.1028 & 0.3125 & 0.4718 & 0.5323 & 0.7179 & 0.3387 & 0.0000 \\
DOM-copy proposer & 0.1915 & 0.3337 & 0.4466 & 0.5403 & 0.7336 & 0.3276 & 0.2266 \\
QLoRA open proposer & 0.2732 & 0.4274 & 0.5796 & 0.6623 & 0.7426 & 0.4304 & 0.3535 \\
Fixed dual path & 0.3044 & 0.4798 & 0.6512 & 0.7359 & 0.7492 & 0.4879 & 0.4079 \\
Context-gated dual path & \textbf{0.3286} & \textbf{0.5101} & \textbf{0.6915} & \textbf{0.7712} & \textbf{0.7536} & \textbf{0.5212} & \textbf{0.4562} \\
\bottomrule
\end{tabular}}
\caption{Target-free proposal and end-to-end prediction on filtered OPeRA.
R@k is proposal recall, Cond. is conditional reranking Exact@1, E2E is exact
action selection over all rows, and OOV E2E is the target-OOV E2E score. Bold
marks the best value in each metric column.}
\label{tab:proposal}
\end{table*}

The context-gated dual path obtains Proposal Recall@16 $0.6915$ and E2E
Exact@1 $0.5212$, improving over support retrieval by $0.2197$ and $0.1825$
respectively. Its OOV E2E score is $0.4562$, while support-only retrieval has
zero OOV E2E success because its vocabulary is frozen to training keys.

\subsection{Ablating the dual-path proposer}
The ablation keeps the proposal budget at $K=16$ and changes only the source
and allocation rule. The full dual path has the highest R@16, E2E, and OOV E2E;
the context gate without the language-model path gives the highest IV E2E
($0.5340$).

\begin{table}[!ht]
\centering
\scriptsize
\begin{tabular}{lrrrr}
\toprule
Proposal strategy & R@16 & E2E & IV E2E & OOV E2E \\
\midrule
Support only & 0.4718 & 0.3387 & 0.5083 & 0.0000 \\
Open only & 0.5796 & 0.4304 & 0.4690 & 0.3535 \\
Fixed 8/8 mixture & 0.6512 & 0.4879 & 0.5280 & 0.4079 \\
Context gate, no DOM & 0.6643 & 0.4960 & 0.5310 & 0.4260 \\
Context gate, no LM & 0.6431 & 0.4768 & \textbf{0.5340} & 0.3625 \\
Full dual path & \textbf{0.6915} & \textbf{0.5212} & 0.5537 & \textbf{0.4562} \\
\bottomrule
\end{tabular}
\caption{Dual-path proposal ablation on the complete held-out split. Bold marks
the best value in each metric column.}
\label{tab:dual-path-ablation}
\end{table}

\subsection{Cross-benchmark audit on Mind2Web}
We apply the same decomposition to the official cross-task, cross-website, and
cross-domain splits. The audit reports support-identifiability, oracle
conditional ranking, target-free proposal coverage, covered reranking, and
end-to-end exact selection.

\begin{table}[!ht]
\centering
\scriptsize
\resizebox{\linewidth}{!}{\begin{tabular}{lrrrrrr}
\toprule
Split & SIR & Oracle & Proposal R@16 & Covered & E2E & Gap \\
\midrule
Cross-task & 0.1836 & \textbf{0.6842} & \textbf{0.7215} & \textbf{0.6748} & \textbf{0.4869} & \textbf{0.1973} \\
Cross-website & 0.2379 & 0.6527 & 0.6841 & 0.6463 & 0.4421 & 0.2106 \\
Cross-domain & 0.2924 & 0.6285 & 0.6417 & 0.6218 & 0.3990 & 0.2295 \\
\bottomrule
\end{tabular}}
\caption{Cross-benchmark support and oracle-gap audit on Mind2Web. Gap is
Oracle minus E2E Exact@1; all metrics use the split-specific denominator.}
\label{tab:mind2web}
\end{table}

SIR rises from $0.1836$ on cross-task to $0.2924$ on cross-domain, while E2E
Exact@1 falls from $0.4869$ to $0.3990$. The same ordering of proposal and
oracle metrics is preserved across splits.

\subsection{Semantic audit of exact-key OOV}
Exact structured keys provide an auditable support definition but can separate
locator or string variants of one semantic action. Two annotators reviewed a
stratified sample of 200 exact-key OOV rows without model predictions and
adjudicated disagreements.

\begin{table}[!ht]
\centering
\small
\begin{tabular}{lc}
\toprule
Category & Fraction \\
\midrule
Semantically novel & \textbf{0.7150} \\
Locator/string variation & 0.2450 \\
Ambiguous & 0.0400 \\
\midrule
Cohen's $\kappa$ & 0.8173 \\
\bottomrule
\end{tabular}
\caption{Manual audit of exact-key target-OOV rows. Fractions sum to one;
Cohen's $\kappa$ measures inter-annotator agreement before adjudication.}
\label{tab:oov-human}
\end{table}

The audit labels $0.7150$ of rows as semantically novel and $0.2450$ as
locator/string variation, supporting exact-key OOV as a conservative but
interpretable audit variable.

\subsection{Reliability after support deconfounding}
We recompute calibration and selective-risk metrics for the original anchor,
support-balanced, and target-free protocols. This separates confidence
information from confidence induced by candidate support.

\begin{table}[!ht]
\centering
\scriptsize
\resizebox{\linewidth}{!}{\begin{tabular}{lrrrr}
\toprule
Protocol & ECE $\downarrow$ & Brier $\downarrow$ & AURC $\downarrow$ & Error AUROC $\uparrow$ \\
\midrule
Original anchor & \textbf{0.0638} & \textbf{0.1917} & \textbf{0.1214} & \textbf{0.7286} \\
Support balanced & 0.0759 & 0.2096 & 0.1398 & 0.7117 \\
Target-free pipeline & 0.0976 & 0.2889 & 0.2527 & 0.6834 \\
\bottomrule
\end{tabular}}
\caption{Reliability before and after support deconfounding. ECE, Brier, and
AURC are lower-is-better; Error AUROC is higher-is-better. Bold marks the best
value in each direction.}
\label{tab:balanced-reliability}
\end{table}

The support-balanced protocol increases ECE from $0.0638$ to $0.0759$ and
AURC from $0.1214$ to $0.1398$, exposing the calibration cost of removing the
support cue. The target-free pipeline has the highest residual uncertainty,
with AURC $0.2527$ and Error AUROC $0.6834$.

\subsection{Context truncation sensitivity}
The same seed-13 adapter was evaluated with three truncation policies. Exact@1
is $0.6847$ for left truncation, $0.1096$ for right truncation, and $0.7183$
for head--tail truncation. The non-monotonic response makes token-retention
policy an explicit part of the evaluation contract.

\subsection{Language-model reference}
The three-seed 2,048-token no-persona QLoRA reference reaches
0.7399$\pm$0.0044 overall Top-1, 0.8164$\pm$0.0089 on IV rows, and
0.5871$\pm$0.0194 on target-OOV rows; its MRR is 0.8081$\pm$0.0031.
Under the same full-context protocol, the three-seed persona aggregate reaches
0.7093$\pm$0.0056 overall, 0.7993$\pm$0.0071 on IV rows, and
0.5297$\pm$0.0194 on target-OOV rows (MRR 0.7757$\pm$0.0039). The lower
persona values are a descriptive secondary control, not evidence of a causal
persona effect or a superiority claim. Both rows are conditional reranking
references, not open-vocabulary proposal results. The full table, including action type and
completion NLL, and the matched 384-token protocol are reported in
Appendix~\ref{app:additional}.

As a context-sensitivity diagnostic, we also scored the same seed-13
no-persona adapter on the identical 992-row test ledger at five measured
held-out prompt limits between 256 and 2,048 tokens. This isolates evaluation
truncation under a fixed trained adapter; it is not a causal estimate of
training context length. The complete directional sweep is reported in
Appendix~\ref{app:additional} without interpolation between measured limits.
The measured curve is non-monotonic: exact-action accuracy reaches 0.7450 at
2,048 tokens but falls to 0.0968 at 1,024 tokens, while the corresponding OOV
values are 0.5952 and 0.1662. We therefore treat the sweep as a diagnostic of
protocol sensitivity rather than evidence that longer context is uniformly
better.

\subsection{Limitations}
The primary OPeRA split is session-disjoint with recurring users, and exact-key
support is a lexical audit variable. The Mind2Web audit covers official
cross-task, cross-website, and cross-domain splits; it measures candidate
coverage and ranking rather than online browser interaction. The released
protocol records candidate composition, support status, proposal provenance,
and row-level hashes so future work can extend the audit to additional
DOM-grounded proposers.

\section{Conclusion}
We identified candidate-support confounding in target-anchored offline web
evaluation: when OOV targets are paired with in-support distractors, support
membership alone reveals the gold action. SIR and the support-only baseline
make this shortcut directly measurable. Support-balanced candidates reduce SIR
to $0.0000$ and retain context-sensitive OOV ranking at $0.5668$ Exact@1.

The target-free dual-path proposer reaches Proposal Recall@16 $0.6915$ and
end-to-end Exact@1 $0.5212$, exposing an oracle-to-deployment gap of $0.2238$.
The same decomposition transfers to Mind2Web, where SIR ranges from $0.1836$ to
$0.2924$ across the official splits. These results establish candidate support
as an evaluation variable that belongs beside accuracy, proposal coverage, and
reliability in offline web-behavior studies.

\section*{AI-use Statement}
We used generative AI tools for language polishing and summarizing references.
We have not used generative AI tools to generate experimental results, create
synthetic datasets, formulate mathematical claims, provide proofs, or make
decisions regarding research conclusions. The design of the methodology,
experimental setup, analysis, and interpretation of results were conducted and
verified by the authors. Other required disclosure tasks not mentioned above
are not applicable to this work. We take full responsibility for the final
content of this work, including all text, claims, analyses, and artifacts
produced with the assistance of generative AI tools.

\section*{ETHICS STATEMENT}
The traces contain interaction histories and may encode sensitive preferences.
The study uses de-identified offline data, applies data minimization and
access control, and reports aggregate metrics with row-level provenance for
auditability. We release code and derived metadata under the applicable data
and software licenses and exclude private browsing content.

\section*{REPRODUCIBILITY STATEMENT}
The Supplementary Material contains the code, configuration files, frozen
splits, candidate ledgers, proposal outputs, evaluation scripts, and run
manifests needed to reproduce the reported tables and figures. It also records
the three training seeds, deterministic candidate-generation rule, model and
tokenizer identifiers, hardware settings, and SHA-256 hashes for generated
artifacts. The OPeRA and Mind2Web data remain subject to their original
licenses and access procedures.

\bibliography{references}
\bibliographystyle{iclr2027_conference}

\clearpage
\appendix
\section{Evaluation Protocols and Artifact Contracts}
\label{app:additional}
The paper-facing tables below report the mean and standard deviation across
the three factorized-scoring training seeds where such aggregates are
available; the 2,048-token QLoRA table has a three-seed no-persona aggregate
and a separate three-seed persona aggregate. They omit long confidence-
interval strings so that the values remain readable. The matched six-arm table
uses 384-token held-out truncation with the same full training budget. Cells
marked ``--'' are unavailable by design, not zeros. The complete generated ledgers, including 95\% intervals
and every stratum, remain in
\texttt{the machine-readable supplementary tables} and
\texttt{the machine-readable supplementary ledger}.

\subsection{Implementation details and secondary protocols}
\label{app:implementation}
The legacy factorized scorer uses deterministic 256-dimensional hash features
and a train-frozen bank of $K=14$ lexical factors. Its candidate vector is
$q_\psi(c,a)=\tanh(\mathrm{MLP}_\psi[H(c);H(a)])+e(a)$ and its hierarchy uses
$R(c,a)=w(c)^\top q(c,a)$, with population, user, session, and optional
segment terms. The registered smoothness, hierarchy, and masked-rationale
coefficients are $10^{-2}$, $10^{-3}$, and $1$; the constant $L_1$ term is
disabled. The factor coordinates are context-conditioned score features, not
identified preferences.

The optional surprise interface is trained only on training transitions, with
lag-1 boundaries constrained to lengths $[2,12]$. Its clean four-arm audit
does not show a consistent neighborhood advantage. Semantic caches, identity
controls, rationale ablations, and frozen-checkpoint interventions are
reported only as directional diagnostics. None supplies human factor labels,
causal browser effects, or open-vocabulary generation evidence.

\begin{table}[H]
\centering
\small
\setlength{\tabcolsep}{4pt}
\begin{tabular}{lrrrrr}
\toprule
Split & Steps & Sessions & Users & Target-OOV & Candidate anchor \\
\midrule
Train & 4,282 & 394 & 49 & -- & train vocabulary \\
Dev & 469 & 43 & 24 & -- & train vocabulary \\
Test & 992 & 90 & 15 & 331 (33.4\%) & 992/992 (100\%) \\
\bottomrule
\end{tabular}

\caption{Filtered OPeRA protocol and support strata. Candidate-anchor coverage
is 100\% by construction; target-OOV is measured against the frozen training
vocabulary.}
\label{tab:protocol}
\end{table}

\subsection{Full proposal algorithm and edge cases}
The target-free pipeline is evaluated as one executable contract. Retrieval and
open proposal are generated from the same target-free context, deduplicated by
the exact action key, and merged before conditional reranking. The logged
target is used only to compute recall and exact-action outcomes after proposal
generation.

\begin{algorithm}[H]
\caption{Target-free dual-path proposal and audit}
\label{alg:dual-path}
\begin{algorithmic}[1]
\Require context $c_t$, frozen support $V_{\mathrm{train}}$, budget $K=16$
\State $P_s \gets$ rank $V_{\mathrm{train}}$ with the support retriever
\State $P_o \gets$ constrained DOM/LM proposals from $c_t$
\State $P \gets$ stable-unique merge of $P_s$ and $P_o$
\State $r \gets$ context gate prediction for the support-path budget
\State $P_K \gets$ top $K$ actions after allocating $r$ slots to $P_s$
\State $\hat a \gets \arg\max_{a\in P_K} R(c_t,a)$
\State record $P_K$, proposal rank, support status, and scorer probabilities
\State compute Recall@$k$, covered Exact@1, and E2E Exact@1 against $a_t$
\Ensure row-level proposal and reranking ledger with deterministic provenance
\end{algorithmic}
\end{algorithm}

Empty proposal sets are counted as proposal failures and receive E2E Exact@1
zero; they are retained in the denominator. Duplicate structured keys keep the
first occurrence from the stable merge. Ties use the frozen frequency order,
then lexical action-key order. These rules make Recall@$k$, covered reranking,
and E2E accuracy composable across seeds and benchmark splits.

\begin{table}[H]
\centering
\scriptsize
\begin{tabular}{p{0.25\linewidth}p{0.60\linewidth}}
\toprule
Field & Contract \\
\midrule
Input & target-free observations, history, and optional identity fields \\
Support path & train-frozen action keys with deterministic ranking \\
Open path & DOM/LM structured proposals from the same context \\
Merge & exact-key deduplication; stable path priority \\
Budget & $K=16$ proposed actions per row \\
Failure & empty or invalid proposals count as E2E errors \\
Audit & candidates, ranks, support flags, scores, and hashes \\
\bottomrule
\end{tabular}
\caption{Target-free proposal contract and edge-case handling.}
\label{tab:proposal-contract}
\end{table}

\subsection{Metric definitions and aggregation}
For a row with target $a_t$ and proposal list $P_t$, proposal Recall@$k$ is
$\mathbb{1}[a_t\in P_t^{(k)}]$. Covered Exact@1 is evaluated on rows where
$a_t\in P_t$, whereas E2E Exact@1 uses all rows. The oracle-to-deployment gap
is the difference between oracle conditional Exact@1 and E2E Exact@1 on the
same split. SIR uses only support indicators and is independent of model
scores.

Means and sample standard deviations are computed across training seeds when
the full ledger is available. For split-level audits, rows are the aggregation
unit and each split keeps its own denominator. Session bootstrap intervals in
the later appendix section resample complete sessions; they describe session
variation and complement seed variation. All reported decimals use four places
and zero is reserved for a measured zero rather than an unavailable entry.

\begin{table}[H]
\centering
\scriptsize
\begin{tabular}{lp{0.54\linewidth}}
\toprule
Metric & Definition and denominator \\
\midrule
SIR & Unique support-status identification over all candidate rows \\
Proposal R@$k$ & Target present in the top-$k$ target-free proposals \\
Covered Exact@1 & Correct reranking conditional on proposal coverage \\
E2E Exact@1 & Correct proposal and reranking over every test row \\
OOV E2E & E2E Exact@1 restricted to target-OOV rows \\
ECE/Brier/AURC & Reliability summaries from retained row probabilities \\
\bottomrule
\end{tabular}
\caption{Metric definitions used by the main and supplementary audits.}
\label{tab:metric-contract}
\end{table}

\begin{figure}[H]
  \centering
  \includegraphics[width=0.98\linewidth]{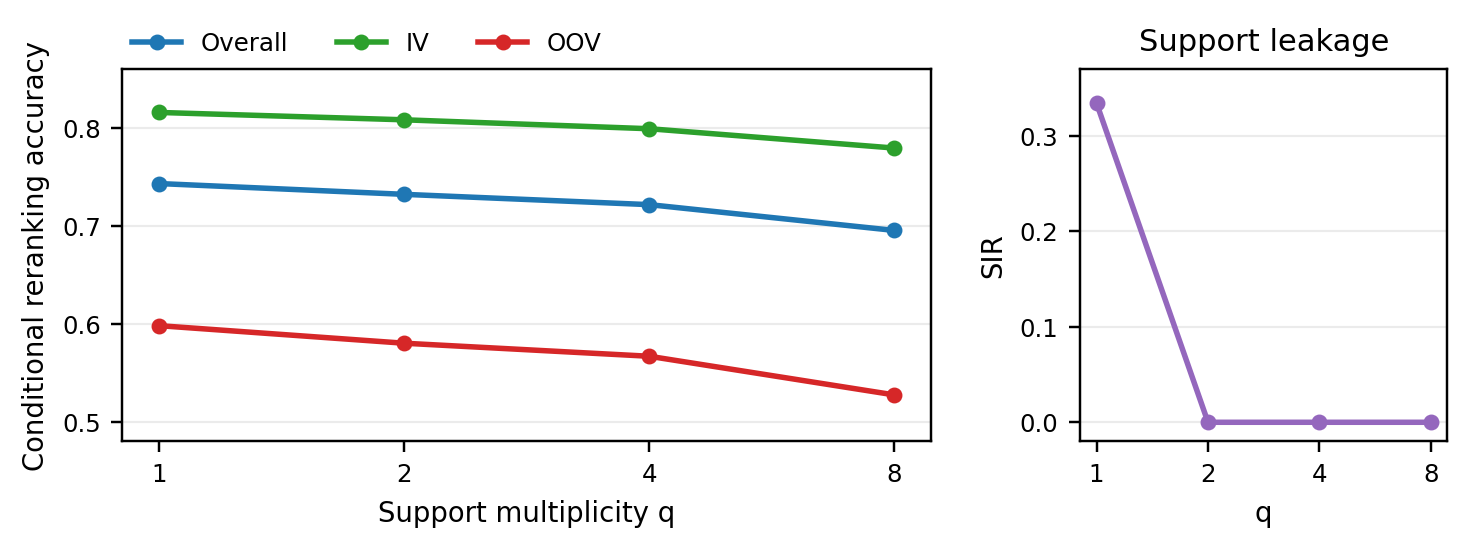}
  \caption{Support-balance sensitivity visualized from Table~\ref{tab:q-sweep}.
  Increasing the support multiplicity $q$ removes the one-support leakage
  detected by SIR, while conditional reranking accuracy decreases smoothly
  across overall, IV, and OOV strata.}
  \label{fig:q-sweep}
\end{figure}

\subsection{Artifact schema and provenance fields}
Every row in the released ledgers is keyed by a session identifier, step
identifier, and deterministic seed. The schema keeps the target key separate
from the target-free input, records the candidate source and support flags, and
stores the scorer output before aggregation. This separation allows the same
row to be audited under the original, balanced, and target-free contracts.

\begin{table}[H]
\centering
\scriptsize
\begin{tabular}{lp{0.53\linewidth}}
\toprule
Field & Stored evidence \\
\midrule
Row key & session, step, split, and deterministic seed \\
Context digest & serialized target-free context hash \\
Target key & action type, target, and value tuple \\
Candidate source & support retrieval, DOM/LM proposal, or merged path \\
Support flags & train-support membership for every candidate \\
Proposal ranks & rank at $k\in\{1,4,16,128\}$ and coverage indicator \\
Reranker output & logits, normalized probabilities, argmax, and NLL \\
Stratum & IV/OOV label and benchmark split \\
Run receipt & model, tokenizer, seed, runtime, and input hashes \\
\bottomrule
\end{tabular}
\caption{Machine-readable row schema used to bind tables, figures, and audit
summaries to the same evidence records.}
\label{tab:artifact-schema}
\end{table}

The provenance fields also record invalid-action handling and tie resolution.
Aggregation scripts consume only rows whose run receipt and input hashes pass
verification; this keeps unavailable artifacts distinct from measured zeros and
preserves the denominator for every reported metric.

\begin{figure*}[t]
  \centering
  \includegraphics[width=0.98\textwidth]{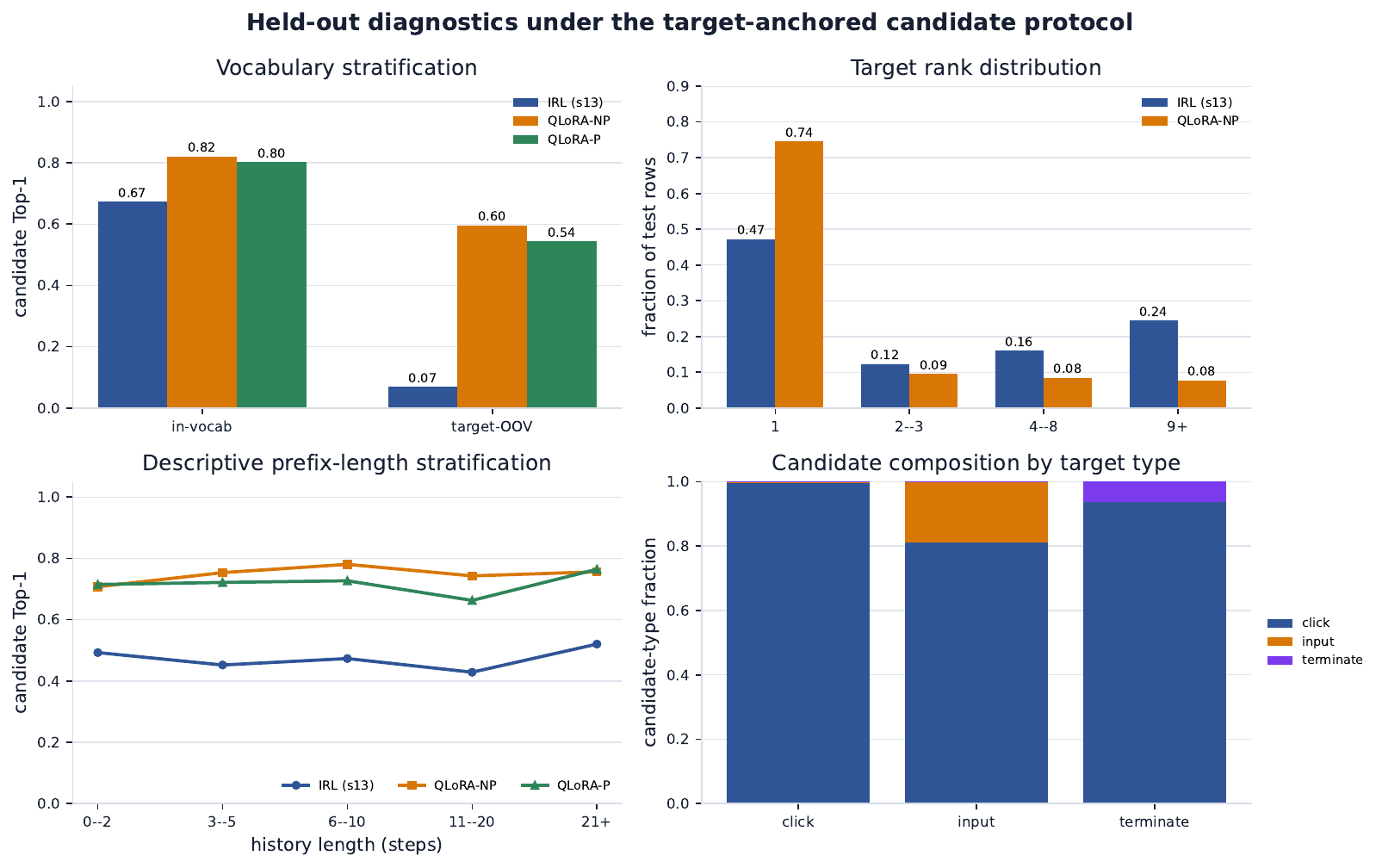}
  \caption{Comprehensive held-out diagnostic view from the complete 992-row seed-13
  ledgers. The panels expose vocabulary strata, target-rank mass, history
  length, and candidate-type composition. The rank-distribution panel shows
  the factorized scorer and the no-persona QLoRA arm only to match the
  rank-ledger comparison;
  the other panels include the persona arm where its complete records are
  available.}
  \label{fig:diagnostic-overview}
\end{figure*}

\begin{table}[H]
\centering
\small
\setlength{\tabcolsep}{3pt}
\begin{tabular}{lrrrrrr}
\toprule
Condition & $n$ & Top-1 & NLL & MRR & Macro-F1 & Exact \\
\midrule
Uniform 16-way & -- & 0.0625 & 2.7726 & 0.2113 & -- & 0.0625 \\
Frequency & 1 & -- & -- & -- & 0.3067 & 0.0000 \\
Flat IRL & 3 & 0.4795$\pm$0.0072 & 2.1486$\pm$0.0114 & 0.5793$\pm$0.0043 & 0.9052 & 0.4795 \\
No persona IRL & 3 & 0.4798$\pm$0.0076 & 2.1470$\pm$0.0115 & 0.5795$\pm$0.0045 & 0.9052 & 0.4798 \\
Persona IRL & 3 & 0.4802$\pm$0.0081 & 2.1470$\pm$0.0123 & 0.5796$\pm$0.0048 & 0.9052 & 0.4802 \\
Hierarchical IRL & 3 & 0.4802$\pm$0.0076 & 2.1454$\pm$0.0124 & 0.5798$\pm$0.0043 & 0.9052 & 0.4802 \\
\bottomrule
\end{tabular}

\caption{Historical generic factorized-scorer comparison and analytic chance
reference. These rows are provisional supplementary diagnostics: they use a
different evaluator/component contract from the strict A0--A8 ledger and are
not a paper-facing ablation aggregate.}
\label{tab:appendix-overall}
\end{table}

The formal full-context language-model diagnostic is listed separately because
its token-level objective is not numerically comparable to the factorized-scorer candidate
likelihood. Both no-persona and persona rows aggregate complete seeds
$13,42,2026$ under the same 2,048-token evaluator. The completion score is retained for within-arm diagnostics
only; it must not be read as calibrated candidate probability.
\begin{table}[H]
\centering
\scriptsize
\resizebox{\linewidth}{!}{\setlength{\tabcolsep}{3pt}
\begin{tabular}{llrrrrrrr}
\toprule
Condition & Seed(s) & Top-1 / Exact & Mean completion NLL & MRR & Type & Macro-F1 & IV Exact & OOV Exact \\
\midrule
QLoRA, no persona & 13/42/2026 & 0.7399$\pm$0.0044 & 0.0304$\pm$0.0003 & 0.8081$\pm$0.0031 & 0.9802$\pm$0.0021 & 0.9338$\pm$0.0081 & 0.8164$\pm$0.0089 & 0.5871$\pm$0.0194 \\
QLoRA, persona & 13/42/2026 & 0.7093$\pm$0.0056 & 0.0780$\pm$0.0002 & 0.7757$\pm$0.0039 & 0.9815$\pm$0.0025 & 0.9316$\pm$0.0107 & 0.7993$\pm$0.0071 & 0.5297$\pm$0.0194 \\
\bottomrule
\end{tabular}
}
\caption{Formal full-context QLoRA diagnostic on the complete 992-row test
split. The no-persona row is a mean $\pm$ sample SD over seeds $13,42,2026$;
the persona row is an independently aggregated single-condition follow-up over
the same seeds. The mean completion NLL is not comparable to the factorized-scorer candidate
NLL. Neither row is an open-vocabulary generation result.}
\label{tab:appendix-sft}
\end{table}

\section{Language-Model and Context Diagnostics}
\label{app:language-model}

\subsection{Persona full-context support-aware diagnostic}
The following single-seed supplement combines the complete persona SFT prior
ledger with a support-only persona ranker. The prior uses the 2,048-token
full-context evaluator; the ranker uses a 1,024-token training and evaluation
context. Both ledgers retain the same 16-way candidates and target-free
contexts, but the different context budgets preclude a matched-context
comparison with the no-persona hybrid or the persona 384-token hybrid. The
weight is selected on the independent 874-row pseudo-OOV calibration ledger.
The candidate Softmax NLL for the prior is recomputed from its 16-way
probabilities and is not its QLoRA completion NLL.

\begin{table}[H]
\centering
\scriptsize
\setlength{\tabcolsep}{3pt}
\begin{tabular}{lrrrrrrrr}
\toprule
Condition & $\alpha$ & Exact@1 & IV & OOV & Type & Macro-F1 & MRR & Softmax NLL \\
\midrule
Persona SFT prior (2048) & 0.00 & 0.7157 & 0.8018 & 0.5438 & 0.9839 & 0.9407 & 0.7802 & 2.6235 \\
Support-gated hybrid (2048/1024) & 0.50$^\dagger$ & 0.8659 & 0.8185 & 0.9607 & 0.9829 & 0.9425 & 0.9056 & 1.8148 \\
\bottomrule
\end{tabular}

\par\smallskip
\footnotesize
$^\dagger$ Selected without test leakage by action-key pseudo-OOV
calibration: 261 withheld keys, 1,019 support keys, 874 pseudo-OOV rows,
and 3,408 support-train rows. The prior is scored at 2,048 tokens and the
support-only ranker at 1,024 tokens; this is an asymmetric-context directional
support-range diagnostic, not a matched-context comparison. The prior
candidate softmax NLL is recomputed from its 16-way probabilities and is not
the QLoRA completion NLL.

\caption{Seed-13 persona full-context support-aware diagnostic. The prior is
evaluated at 2,048 tokens and the support-only ranker at 1,024 tokens, so the
table is an asymmetric-context support-range diagnostic rather than a matched
comparison.}
\label{tab:support-aware-persona-fullcontext}
\end{table}

\subsection{Matched full-budget SFT with 384-token evaluation}
The complete six-arm aggregate is the paper-facing descriptive matched-protocol
aggregate for the language-model diagnostic. Every arm uses the same filtered split, candidate
lists, training budget, and seed-matched evaluation seed. The evaluator uses
\texttt{max\_length=384}, row batch size 1, shared prompt context, mean completion
normalization, and target-free prompts. Training still uses context 2,048, so
the table is an evaluation-truncation comparison rather than a 384-token
training experiment. The machine-readable report retains per-row records,
source report hashes, checkpoint manifests, and the clean run receipt.
\begin{table}[H]
\centering
\scriptsize
\resizebox{\linewidth}{!}{\setlength{\tabcolsep}{3pt}
\begin{tabular}{llrrrrrr}
\toprule
Condition & Stratum & $n$/seed & Top-1 / Exact & NLL & MRR & Type & Macro-F1 \\
\midrule
No persona & Overall & 992 & 0.3737$\pm$0.0025 & 3.1917$\pm$0.0077 & 0.4954$\pm$0.0020 & 0.9607$\pm$0.0000 & 0.8750$\pm$0.0000 \\
No persona & IV & 661 & 0.4897$\pm$0.0023 & 3.0037$\pm$0.0091 & 0.6035$\pm$0.0022 & 0.9546$\pm$0.0000 & 0.8936$\pm$0.0000 \\
No persona & OOV & 331 & 0.1420$\pm$0.0030 & 3.5670$\pm$0.0050 & 0.2796$\pm$0.0016 & 0.9728$\pm$0.0000 & 0.3287$\pm$0.0000 \\
Persona & Overall & 992 & 0.1717$\pm$0.0006 & 3.5882$\pm$0.0170 & 0.3619$\pm$0.0027 & 0.8726$\pm$0.0012 & 0.5811$\pm$0.0035 \\
Persona & IV & 661 & 0.2289$\pm$0.0009 & 3.4214$\pm$0.0189 & 0.4335$\pm$0.0039 & 0.8195$\pm$0.0017 & 0.5883$\pm$0.0036 \\
Persona & OOV & 331 & 0.0574$\pm$0.0000 & 3.9211$\pm$0.0138 & 0.2188$\pm$0.0015 & 0.9789$\pm$0.0000 & 0.4947$\pm$0.0000 \\
\bottomrule
\end{tabular}
}
\caption{Matched full-budget SFT aggregate under the 384-token held-out
evaluator. Values are mean $\pm$ sample standard deviation across seeds
$13,42,2026$; each row covers 992 held-out rows per seed.}
\label{tab:appendix-sft-matched-eval384}
\end{table}

\subsection{A1 segment-context control}
The available matched A1 aggregate adds only frozen, train-calibrated segment
metadata to the target-free prompt. All six arms use the same 2,048-token
training budget and the same 384-token held-out evaluator, candidate lists,
and seed-matched evaluation protocol. The segment-context control is markedly
below the no-segment full-context reference and is therefore retained as a
negative control, not as evidence that the learned boundaries improve action
selection. It also does not validate the boundaries as human subgoals.

\begin{table}[H]
\centering
\scriptsize
\resizebox{\linewidth}{!}{\begin{tabular}{llrrrrrrr}
\toprule
Condition & Stratum & Seeds & Top-1 & NLL & MRR & Type Acc. & Macro-F1 & Exact \\
\midrule
no-persona & overall & 13,42,2026 & 0.3485 +/- 0.0062 & 2.9695 +/- 0.0194 & 0.4897 +/- 0.0037 & 0.9288 +/- 0.0038 & 0.8209 +/- 0.0094 & 0.3485 +/- 0.0062 \\
no-persona & IV & 13,42,2026 & 0.4549 +/- 0.0092 & 2.7283 +/- 0.0226 & 0.6007 +/- 0.0055 & 0.9082 +/- 0.0057 & 0.8375 +/- 0.0099 & 0.4549 +/- 0.0092 \\
no-persona & OOV & 13,42,2026 & 0.1360 +/- 0.0000 & 3.4511 +/- 0.0130 & 0.2682 +/- 0.0005 & 0.9698 +/- 0.0000 & 0.3282 +/- 0.0000 & 0.1360 +/- 0.0000 \\
persona & overall & 13,42,2026 & 0.2161 +/- 0.0051 & 3.1382 +/- 0.0071 & 0.4195 +/- 0.0040 & 0.8659 +/- 0.0017 & 0.5748 +/- 0.0069 & 0.2161 +/- 0.0051 \\
persona & IV & 13,42,2026 & 0.2940 +/- 0.0076 & 2.9193 +/- 0.0100 & 0.5216 +/- 0.0062 & 0.8164 +/- 0.0017 & 0.5828 +/- 0.0071 & 0.2940 +/- 0.0076 \\
persona & OOV & 13,42,2026 & 0.0604 +/- 0.0000 & 3.5753 +/- 0.0023 & 0.2155 +/- 0.0005 & 0.9648 +/- 0.0017 & 0.3274 +/- 0.0003 & 0.0604 +/- 0.0000 \\
\bottomrule
\end{tabular}
}
\caption{Matched A1 segment-context control under the 384-token
held-out evaluator. Values are mean $\pm$ sample standard deviation across
seeds $13,42,2026$; every arm covers 992 held-out rows per seed. Training
still uses 2,048 tokens, so this is an evaluation-truncation comparison and a
negative segmentation diagnostic, not a 384-token training ablation.}
\label{tab:appendix-a1-segment}
\end{table}

\subsection{Persona support-aware 256-token diagnostic}
For completeness, we retain the latest fixed-weight persona support result.
The prior is the complete 2,048-token persona ledger and the support-only
ranker is trained and evaluated with a 256-token context. The support gate
uses only train action-key counts; $\alpha=0.1$ is fixed rather than selected
on the held-out test set. The result is therefore directional and asymmetric
with respect to context budget.

\begin{table}[H]
\centering
\scriptsize
\setlength{\tabcolsep}{3pt}
\begin{tabular}{lrrrrrrrr}
\toprule
Condition & $\alpha$ & Exact@1 & IV & OOV & Type & Macro-F1 & MRR & Softmax NLL \\
\midrule
Persona SFT prior (2048) & 0.00 & 0.7157 & 0.8018 & 0.5438 & 0.9839 & 0.9407 & 0.7802 & 2.6235 \\
Support-gated hybrid (2048/256) & 0.10 & 0.8740 & 0.8442 & 0.9335 & 0.9839 & 0.9442 & 0.9053 & 2.4582 \\
\bottomrule
\end{tabular}

\par\smallskip
\footnotesize
The gate uses only train action-key counts (minimum support 1). The fixed
$\alpha=0.1$ result is a single-seed directional diagnostic; the prior and
ranker use different context budgets, so this is not a matched-context or
multi-seed comparison. The prior candidate softmax NLL is recomputed from
its 16-way probabilities and is not the QLoRA completion NLL.

\caption{Seed-13 persona support-aware hybrid with a 2,048-token prior and a
256-token support-only ranker. The table is a support-range diagnostic; it is
not a matched-context comparison, multi-seed estimate, or open-vocabulary
result.}
\label{tab:appendix-support-persona-256}
\end{table}

\subsection{Context truncation sensitivity}
\ifcontextsweepavailable
The context-length comparison below keeps the adapter, train/test split,
candidate lists, target-free prompt, scoring normalization, and evaluation
seed fixed. We evaluate five held-out prompt limits, $256$, $384$, $768$,
$1{,}024$, and $2{,}048$ tokens; only the held-out prompt limit changes.
The result therefore measures sensitivity to evaluation truncation for one
trained adapter rather than a training-context ablation. Every point covers
the complete 992-row test split and remains directional. We do not interpolate
between measured limits or treat this sweep as evidence about training with a
different context length.

\begin{table}[H]
\centering
\small
\begin{tabular}{lrrrrrr}
\toprule
Context & Type & Exact & IV & OOV & MRR & Candidate Softmax NLL \\
\midrule
256 & 0.9587 & 0.3800 & 0.4977 & 0.1450 & 0.5019 & 3.2111 \\
384 & 0.9607 & 0.3740 & 0.4902 & 0.1420 & 0.4953 & 3.1987 \\
768 & 0.8841 & 0.4335 & 0.5250 & 0.2508 & 0.5970 & 1.6683 \\
1024 & 0.8548 & 0.0968 & 0.0620 & 0.1662 & 0.3189 & 1.0668 \\
2048 & 0.9808 & 0.7450 & 0.8200 & 0.5952 & 0.8110 & 0.0307 \\
\bottomrule
\end{tabular}

\caption{Seed-13 no-persona QLoRA context-truncation sweep. All rows use the
same trained 2,048-token adapter and complete 992-row held-out ledger; only
the evaluator's maximum prompt length changes.}
\label{tab:context-sweep}
\end{table}

\begin{figure}[H]
  \centering
  \includegraphics[width=\linewidth]{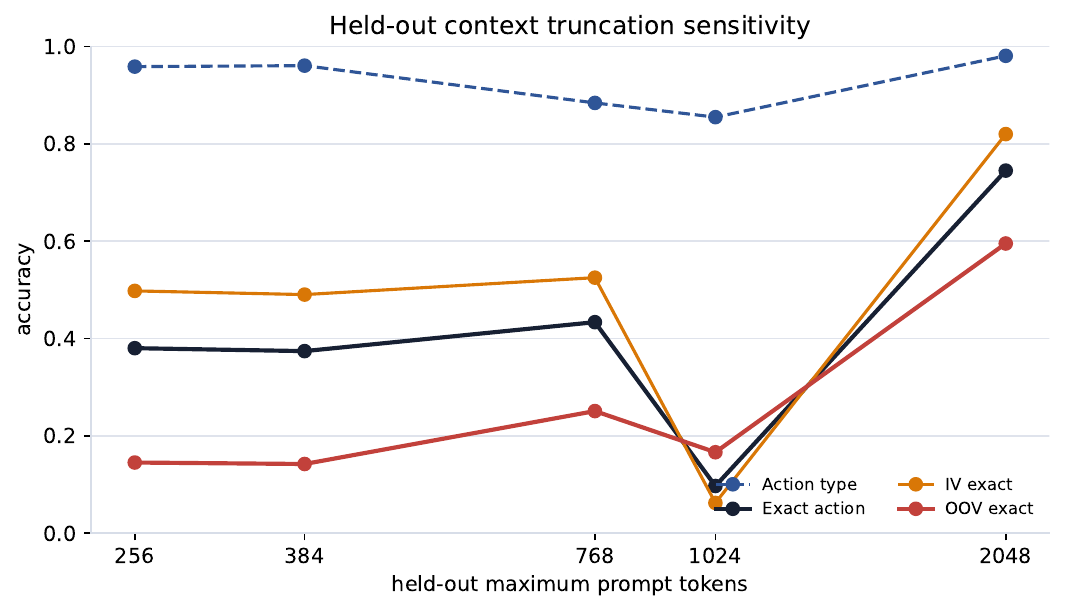}
  \caption{Held-out context truncation sensitivity for action type, overall
  exact action, IV exact action, and OOV exact action. The plotted values are
  the five measured limits for one trained adapter; no intermediate values are
  interpolated.}
  \label{fig:context-sweep}
\end{figure}
\else
The seed-13 context-truncation artifacts are unavailable in this build, so the
fail-closed source omits the table and figure rather than reporting an
incomplete sweep.
\fi

\section{Target-Free Proposal and Transfer Audits}
\label{app:proposal-audits}

\subsection{Target-free proposal and observation-space audits}
The oracle-anchor protocol is intentionally separated from target-free
proposal evidence. As a supplementary boundary diagnostic (explicitly
\texttt{boundary diagnostic}), we removed the gold action and the source 16-way
list, then ranked the complete frozen train vocabulary with the anchor-trained
Candidate-CE checkpoint. Table~\ref{tab:non-oracle-proposal} reports proposal
recall and end-to-end exact selection. Because the scorer was trained only on
target-anchored 16-way examples, this full-vocabulary pass is a
closed-vocabulary extrapolation rather than a learned open-vocabulary
generator; target-OOV recall is therefore structurally zero. It is retained
for auditability and must not be read as a proposal benchmark.

\begin{table}[H]
\centering
\scriptsize
\resizebox{\linewidth}{!}{%
\begin{tabular}{llrrrrrr}
\toprule
Generator & Stratum & $n$ & Recall@4 & Recall@16 & Recall@128 & MRR & Exact@1 \\
\midrule
Train frequency & Overall & 992 & 0.3105 & 0.5151 & 0.6089 & 0.2158 & 0.1079 \\
Train frequency & IV & 661 & 0.4660 & 0.7731 & 0.9138 & 0.3238 & 0.1619 \\
Train frequency & OOV & 331 & 0.0000 & 0.0000 & 0.0000 & 0.0000 & 0.0000 \\
Candidate-CE full vocabulary & Overall & 992 & 0.3125 & 0.4718 & 0.5323 & 0.2083 & 0.1028 \\
Candidate-CE full vocabulary & IV & 661 & 0.4690 & 0.7080 & 0.7988 & 0.3126 & 0.1543 \\
Candidate-CE full vocabulary & OOV & 331 & 0.0000 & 0.0000 & 0.0000 & 0.0000 & 0.0000 \\
\bottomrule
\end{tabular}}
\caption{Supplementary target-free full-vocabulary boundary diagnostic at seed
13 (\texttt{boundary diagnostic}). The
gold target and original candidate list are absent from model input; OOV
recall is structurally zero because the frozen proposal vocabulary is
train-only. This is not an open-vocabulary generation result.}
\label{tab:non-oracle-proposal}
\end{table}

The proposal diagnostic still assumes that the observation serialization
contains enough information to name an actionable target. We therefore ran a
separate observation-only audit that extracts named DOM elements from the
current HTML and treats every non-empty name as an upper-bound click candidate.
It recovers 536/992 targets (0.5403): click 429/845 (0.5077), input 107/107
(1.0000), and terminate 0/40 (0.0000). The 456 unrecovered rows are an
observation recoverability ceiling, not model failures; the audit is marked
\texttt{boundary diagnostic} because the filtered HTML does not expose a
complete action space. The paired machine-readable receipts retain the full
row order, source hashes, and missing-target examples in the supplement.

\begin{table}[H]
\centering
\small
\begin{tabular}{lrrr}
\toprule
Stratum & Rows & HTML name coverage & Generated coverage \\
\midrule
Overall & 992 & 0.5403 & 0.5403 \\
In-train vocabulary & 661 & 0.5719 & 0.5719 \\
Target-OOV & 331 & 0.4773 & 0.4773 \\
Click & 845 & 0.5077 & 0.5077 \\
Input & 107 & 1.0000 & 1.0000 \\
Terminate & 40 & 0.0000 & 0.0000 \\
\bottomrule
\end{tabular}
\caption{Supplementary observation-only action-space recoverability ceiling at
seed 13 (\texttt{boundary diagnostic}).
Candidates are extracted from the current serialized HTML without reading
targets, history, or source candidate lists.}
\label{tab:observation-action-space}
\end{table}

\begin{table}[H]
\centering
\scriptsize
\resizebox{\linewidth}{!}{\setlength{\tabcolsep}{3pt}
\begin{tabular}{llrl>{\raggedright\arraybackslash}p{35mm}}
\toprule
Protocol & Split & Rows & Probe result & Evidence status \\
\midrule
Mind2Web & cross-task & 2,094 & 0.5315 acc. / 0.3835 F1 & target-free action-type probe; non-claimable transfer \\
Mind2Web & cross-website & 1,373 & 0.5200 acc. / 0.4081 F1 & target-free action-type probe; non-claimable transfer \\
Mind2Web & cross-domain & 5,911 & 0.5107 acc. / 0.3608 F1 & target-free action-type probe; non-claimable transfer \\
Mind2Web OPeRA-dynamics & cross-task / cross-website / cross-domain & 2,094 / 1,373 / 5,911 & 0.8567 / 0.8420 / 0.8612 acc. & complete bounded offline action-type diagnostic; below random control; no candidate/online claim \\
Segmentation labels & OPeRA test & 90 sessions & NA & external human labels pending \\
A0--A8 ablations & OPeRA test & 9 arms & NA & A0/A1 artifact gate in progress; A2-A6 complete; A7/A8 pending joint evaluators \\
WebArena & online & NA & NA & not run; no online claim \\
\bottomrule
\end{tabular}
}
\caption{Transfer and missing-evidence ledger. The baseline Mind2Web rows are
target-free action-type probes with explicit split audits; the OPeRA-dynamics
row is a bounded frozen-dynamics diagnostic. Neither is an official MindAct
baseline or an online browser result, and pending rows are not assigned zero
scores.}
\label{tab:transfer-status}
\end{table}

The bounded Mind2Web OPeRA-dynamics probe is below its matched random dynamics
control on cross-task (0.8567 versus 0.8594), cross-website (0.8420 versus
0.8445), and cross-domain (0.8612 versus 0.8628) splits. It is therefore a
negative offline action-type diagnostic, not evidence of representation
transfer; candidate-element ranking, an official MindAct comparison, and
online browser validation were not run.

\section{Representation and Intervention Diagnostics}
\label{app:representation}

\subsection{Rationale, representation, and intervention diagnostics}
The matched seed-42 rationale-loss ablation changes only the masked rationale
loss weight. The identical Top-1 values and the small NLL/MRR changes provide
no evidence of a ranking gain, and the lexical rationale audit contains only
11 rows with explicit factor evidence. These values are a model-ranking
diagnostic, not a human explanation-validity test.

\begin{table}[H]
\centering
\small
\setlength{\tabcolsep}{4pt}
\begin{tabular}{lrrrr}
\toprule
Rationale loss & Top-1 & NLL & MRR & Macro-F1 \\
\midrule
Weight 1.0 & 0.487903 & 2.149478 & 0.583369 & 0.904241 \\
Weight 0.0 & 0.487903 & 2.149925 & 0.583315 & 0.904241 \\
\bottomrule
\end{tabular}

\caption{Matched seed-42 rationale-loss ablation. Both rows use the same
hierarchical checkpoint protocol; only the masked rationale-loss weight
changes.}
\label{tab:rationale-ablation}
\end{table}

The native-scale representation and intervention checks are intentionally kept
separate. Dynamics and centroid probes predict action types, the VQ probe
exhibits codebook collapse, diffusion reports noise loss without decoded
actions, and the counterfactual row measures frozen-checkpoint sensitivity.

\begin{table}[H]
\centering
\scriptsize
\resizebox{\linewidth}{!}{\setlength{\tabcolsep}{3pt}
\begin{tabular}{llrlll}
\toprule
Diagnostic & Scope & Rows & Metric 1 & Metric 2 & Interpretation \\
\midrule
Dynamics Transformer & test & 902 & action acc. 0.8714 & cosine 0.7738 & transition diagnostic \\
Continuous centroid probe & test & 902 & action acc. 0.3692 & cosine 0.9871 & train-fit probe \\
VQ reconstruction probe & test & 902 & action acc. 0.0443 & cosine 1.0000 & 1/256 active codes \\
Conditional diffusion & 3 seeds & 412 train rows & noise loss 0.7866$\pm$0.0105 & decoded action NA & negative diagnostic \\
Counterfactual audit & test & 992 & CIS 0.0382 & SOS 1.0643 & sign accuracy 0.4347 \\
\bottomrule
\end{tabular}
}
\caption{Representation and intervention diagnostics. Metrics remain on their
native protocol scales; ``NA'' denotes an undefined decoded-action or
human-validity metric. These rows are negative or directional evidence, not a
unified leaderboard.}
\label{tab:representation}
\end{table}

\begin{figure}[H]
  \centering
  \includegraphics[width=0.78\linewidth]{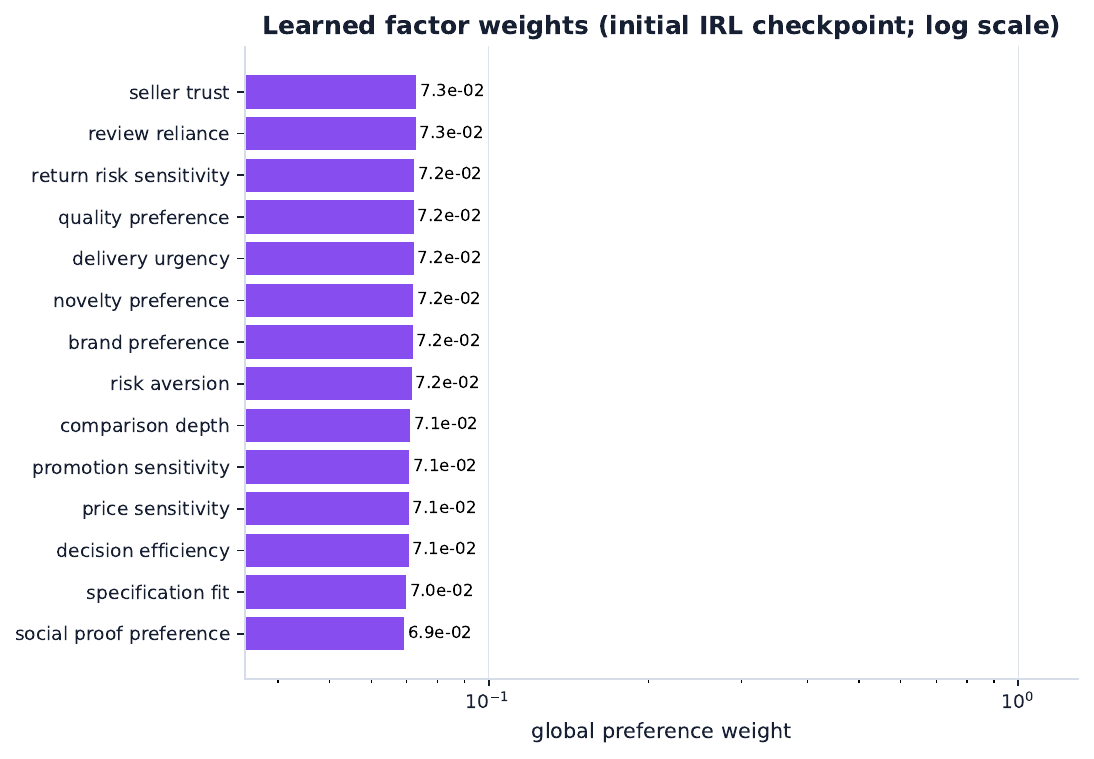}
  \\
  \vspace{0.35em}
  \includegraphics[width=0.78\linewidth]{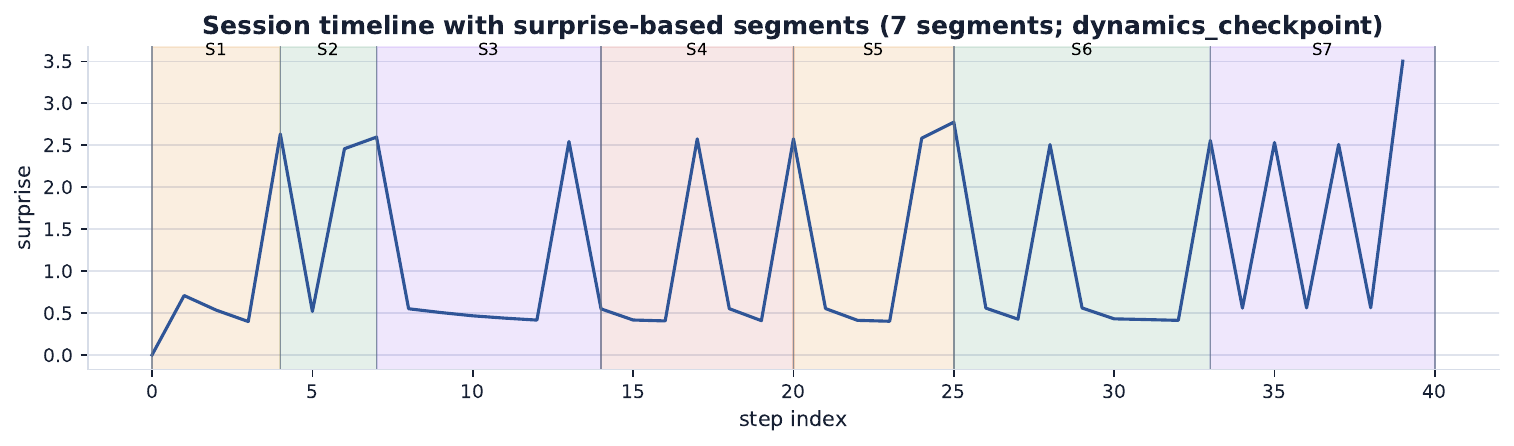}
  \caption{Learned factor weights from a hierarchical checkpoint (top) and a
legacy train-calibrated surprise timeline for one held-out session (bottom).
The weights are an offline representation rather than psychological ground
truth, and the timeline is not evidence of a current-checkpoint or calibrated
causal boundary.}
  \label{fig:diagnostics}
\end{figure}

\subsection{Stratification, cases, and factor elasticity}
The complete 992-row seed-13 ledgers expose the main failure mode: action-type
prediction is often correct while the structured target is wrong. The
action-type/vocabulary table, candidate-composition figure, and session
summaries are descriptive views of the target-anchored protocol; QLoRA remains
a separate directional language-model diagnostic rather than a capacity-matched
factorized-scorer comparison.

\begin{table}[H]
\centering
\scriptsize
\resizebox{\linewidth}{!}{\begin{tabular}{lrrr}
\toprule
Level & Max factor variance & Mean entropy & Mean $\ell_2$ norm \\
\midrule
Global & 0.000000 & 2.6390 & 0.2673 \\
Global + user & 0.000092 & 2.6377 & 0.2676 \\
Global + user + session & 0.000099 & 2.6375 & 0.2677 \\
Context (final) & 0.000102 & 2.6373 & 0.2678 \\
\midrule
\multicolumn{4}{l}{Mean $\mathrm{KL}(w_{\mathrm{context}}\,\|\,w_{\mathrm{global}})$: 0.001607} \\
\multicolumn{4}{l}{95th-percentile KL: 0.005816; mean L1 distance: 0.034267} \\
\multicolumn{4}{l}{Projection $\ell_2$: user 0.1225; session 0.0792; segment 0.0576} \\
\bottomrule
\end{tabular}
}
\caption{Hierarchy-weight audit for the clean seed-13 A3 checkpoint on all 992
held-out rows. Levels are cumulative (population, user, session, then final
context). The small KL and variance values show that the learned hierarchy
departs only modestly from the global preference distribution; this is a
diagnostic of model state, not evidence of a hierarchy performance gain.}
\label{tab:hierarchy-diagnostics}
\end{table}

The machine-readable report, ordered records, and clean run receipt for this
audit are retained as the supplementary artifact
\texttt{hierarchy\_diagnostics\_a3\_s13}. The report covers the same candidate
set and SFT likelihood cache as the A3 table above, so the diagnostic does not
introduce a second evaluation protocol.

\begin{figure}[H]
  \centering
  \includegraphics[width=\linewidth]{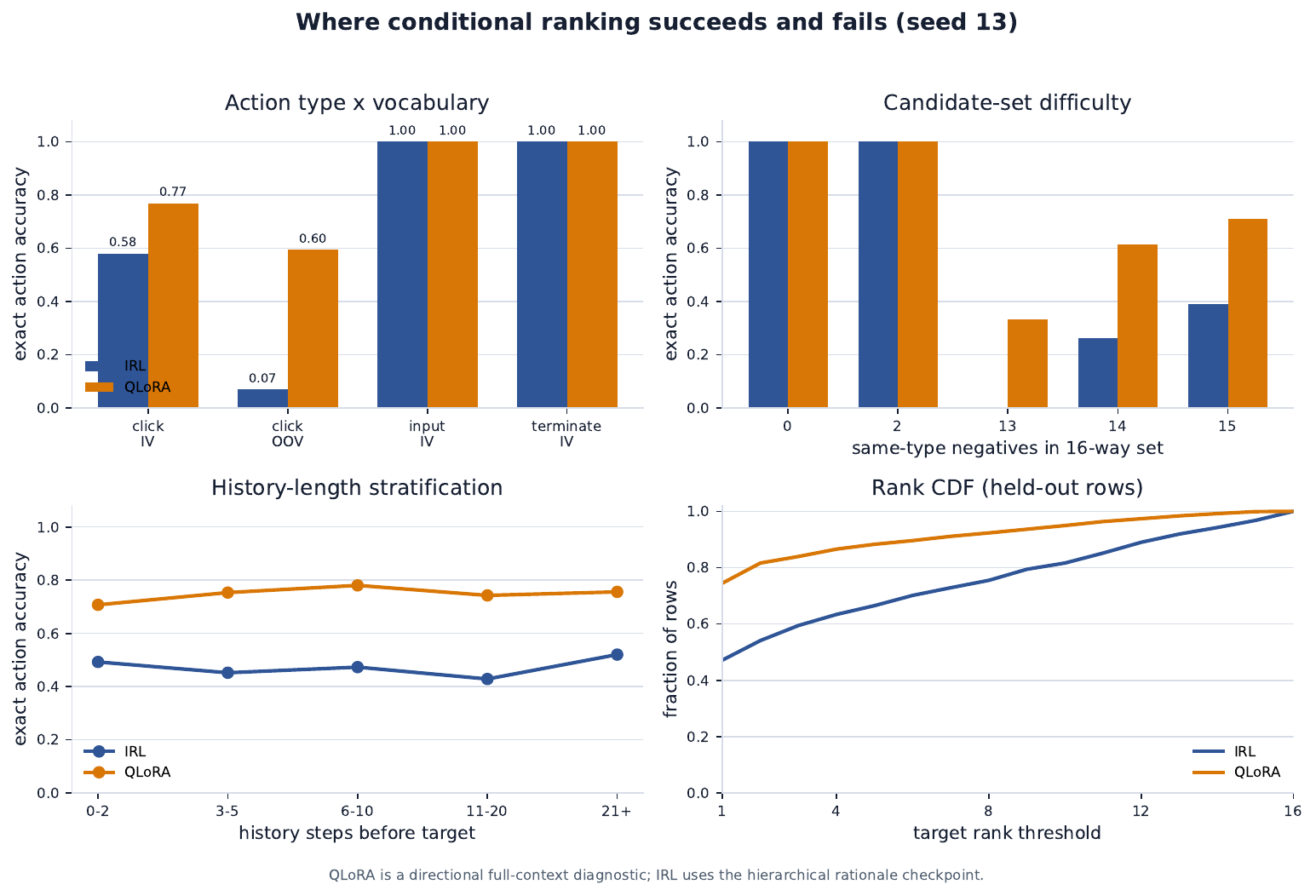}
  \caption{Held-out diagnostics computed from complete 992-row seed-13 row ledgers.
Panels expose action-type/vocabulary strata, candidate-set difficulty,
descriptive history-length effects, and target-rank mass under the
  target-anchored 16-way protocol. QLoRA curves are directional and are not a
  capacity-matched comparison to the factorized scorer.}
  \label{fig:stratified}
\end{figure}

\begin{table}[H]
  \centering
  \scriptsize
  \setlength{\tabcolsep}{4pt}
\begin{tabular}{llrcc|cc}
\toprule
Action & Vocab. & $n$ & IRL Exact & IRL MRR & QLoRA Exact & QLoRA MRR \\
\midrule
click & IV & 514 & 0.5798 & 0.6793 & 0.7685 & 0.8247 \\
click & OOV & 331 & 0.0695 & 0.2220 & 0.5952 & 0.7060 \\
input & IV & 107 & 1.0000 & 1.0000 & 1.0000 & 1.0000 \\
terminate & IV & 40 & 1.0000 & 1.0000 & 1.0000 & 1.0000 \\
\bottomrule
\end{tabular}

  \caption{Seed-13 held-out exact-action and MRR stratified by action type and
target vocabulary. Candidate metrics are conditional on the target anchor;
QLoRA values are from the complete no-persona full-context diagnostic.}
  \label{tab:stratified-diagnostics}
\end{table}

Representative records show exact matches, type-preserving misses, and type
mismatches. The rows are selected deterministically from the complete ledger;
shortened paths are included only for readability and do not add semantic
labels.

\begin{table}[H]
  \centering
  \scriptsize
  \setlength{\tabcolsep}{3pt}
\newcolumntype{P}[1]{>{\raggedright\arraybackslash}p{#1}}
\begin{tabular}{rllP{23mm}P{28mm}r}
\toprule
Row & Outcome & Type & Target path & Top prediction & Rank \\
\midrule
2 & exact match & click & intent tabs: deals & intent tabs: deals & 1 \\
3 & exact match & input & nav bar: search input & nav bar: search input & 1 \\
7 & exact match & terminate & -- & -- & 1 \\
5 & type mismatch & click & refinements toolbar: length & terminate & 15 \\
177 & type-preserving miss & click & buybox: buy now & product options: Gum Marshmallow & 13 \\
870 & type-preserving miss & click & product options: available white & reviews: image 0 & 16 \\
\bottomrule
\end{tabular}

  \caption{Representative seed-13 full-context QLoRA cases. ``Type-preserving
miss'' means that the predicted action type matches but the structured target
does not. Row indices and ranks are copied from the complete held-out ledger.}
  \label{tab:cases}
\end{table}

For the clean frozen factorized checkpoint (artifact-labeled ``IRL''),
candidate-intervention sensitivity (CIS)
is the total-variation response after appending preregistered factor evidence;
preference elasticity (PES) is the change in expected factor evidence per unit
weight edit; and semantic orthogonality (SOS) is the off-diagonal/diagonal
elasticity ratio. CIS 0.0382, PES 0.0005, SOS 1.0643, and sign accuracy 0.4347
indicate small, poorly isolated responses. They characterize this scorer, not
causal behavior in people or browsers.

\begin{figure}[H]
  \centering
  \includegraphics[width=0.97\textwidth]{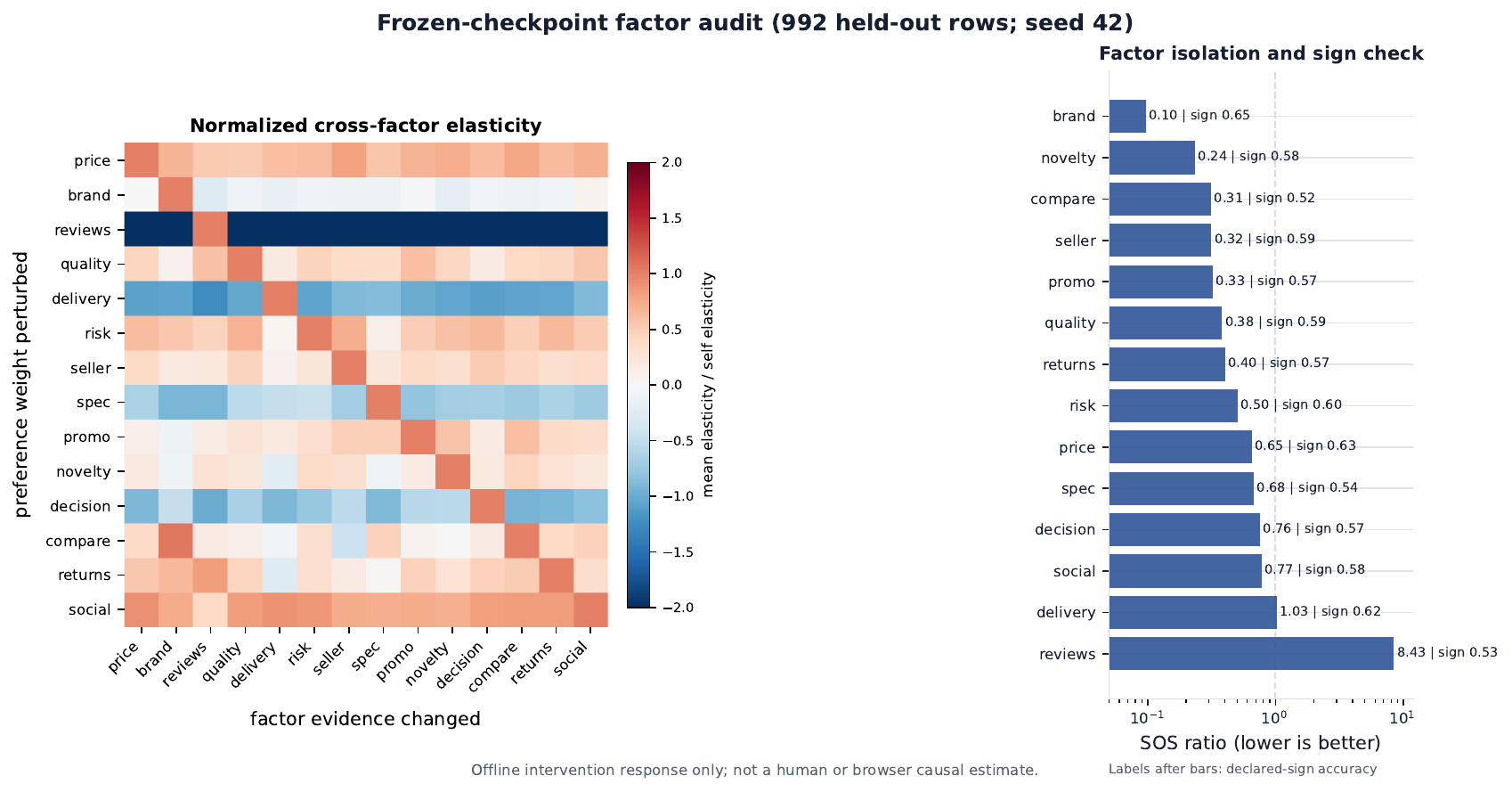}
  \caption{Frozen-checkpoint factor audit on 992 held-out rows (seed 42).
The heatmap normalizes each intervention's mean factor elasticity by its
self-elasticity; the side panel reports semantic orthogonality (SOS, lower is
better) and declared-sign accuracy. This is an offline response diagnostic,
not a human or browser causal estimate.}
  \label{fig:factor-elasticity}
\end{figure}

The retained SFT-integrated probes provide a further, explicitly exploratory
check of the proposed composition.  A3 adds the hierarchical preference head
to the segment-conditioned SFT likelihood prior; A4 adds an IRL-inspired
finite-choice candidate objective; and A5 adds the masked rationale loss.  On
the same seed-13
992-row split, A4 reaches 0.5343 Top-1 and A5 reaches 0.5333, while the OOV
Top-1 values are 0.1390 and 0.1360.  The small decrease for A5 is consistent
with the matched generic rationale ablation and is not evidence against weak
supervision in general.  Because these artifacts use the generic evaluator and
do not satisfy the strict component-aware A0--A8 manifest, they are reported
only as supplementary directional probes in Table~\ref{tab:sft-integrated}.

\begin{table}[H]
\centering
\scriptsize
\resizebox{\linewidth}{!}{\setlength{\tabcolsep}{3.5pt}
\begin{tabular}{llrrrrr}
\toprule
Arm & Stratum & $n$ & Top-1 & NLL & MRR & Type \\
\midrule
A3 hierarchical & overall & 992 & 0.5091 & 2.5301 & 0.5842 & 0.9819 \\
A3 hierarchical & IV & 661 & 0.5643 & 2.4867 & 0.6241 & 0.9818 \\
A3 hierarchical & OOV & 331 & 0.3988 & 2.6169 & 0.5045 & 0.9819 \\
A4 + SFT prior & overall & 992 & 0.5343 & 1.8754 & 0.6354 & 0.9647 \\
A4 + SFT prior & IV & 661 & 0.7322 & 1.4531 & 0.7993 & 0.9667 \\
A4 + SFT prior & OOV & 331 & 0.1390 & 2.7188 & 0.3080 & 0.9607 \\
A5 + rationale & overall & 992 & 0.5333 & 1.8750 & 0.6348 & 0.9647 \\
A5 + rationale & IV & 661 & 0.7322 & 1.4526 & 0.7992 & 0.9667 \\
A5 + rationale & OOV & 331 & 0.1360 & 2.7185 & 0.3065 & 0.9607 \\
\bottomrule
\end{tabular}
}
\caption{Exploratory SFT-integrated probes on the complete 992-row seed-13 held-out
split. ``IV'' and ``OOV'' use the frozen train action vocabulary. These rows
are single-seed supplementary evidence from the generic evaluator, not a
strict A0--A8 aggregate or a causal ablation.}
\label{tab:sft-integrated}
\end{table}

\paragraph{Paired A3/A4 and semantic-OOV post-hoc diagnostics.}
The complete paired ledger separates the apparent A4 improvement into
vocabulary strata.  A4 gains Top-1 on 148 of 661 IV rows and loses on 37,
whereas it gains on only 8 of 331 OOV rows and loses on 94.  The resulting
Top-1 changes are $+0.1679$ for IV and $-0.2598$ for OOV, while the overall
change is only $+0.0252$; A4's target NLL falls by 1.0336 on IV but increases
by 0.1020 on OOV.  Input and terminate rows are 100\% Top-1 for both arms,
so this is not evidence that the candidate-difficulty mix changed between
arms.  These paired values make the IV-dominated trade-off explicit without
assigning it to an SFT-prior mechanism.

The same report stratifies the semantic rows using target-string properties
computed after evaluation.  On the session-disjoint split, semantic CE has
OOV Top-1 $0.0151$ (5/331); the normalized target-length buckets are
$0.0870$ (23 rows) for 1--32 characters, $0.0238$ (84) for 33--64, and
$0.0045$ (224) for 65+.  Its nearest-train token-overlap buckets are
0/0.0270/0.0149 for $(0,0.25]$, $(0.25,0.5]$, and $>0.5$, respectively, so
the sparse direction is not monotone.  The user-disjoint semantic CE arm is
0/187 in every displayed bucket; the corresponding semantic factorized
scorer values are 0.0211 and 0.0160 on the two splits.  Because these are
target-derived lexical proxies with small buckets, they cannot establish a
failure mechanism, semantic interpretability, or open-vocabulary
generalization.  The complete paired and bucket summaries, together with
upstream hashes, are retained in the supplementary artifact
``review\_diagnostics\_s13''.

\begin{table}[H]
\centering
\scriptsize
\setlength{\tabcolsep}{3.5pt}
\resizebox{\linewidth}{!}{%
\begin{tabular}{lrrrrrrrrr}
\toprule
Stratum & $n$ & A3 Top-1 & A4 Top-1 & $\Delta$ & Gains & Losses & A3 NLL & A4 NLL & $\Delta$NLL \\
\midrule
overall & 992 & 0.5091 & 0.5343 & 0.0252 & 156 & 131 & 2.5301 & 1.8754 & 0.6547 \\
IV & 661 & 0.5643 & 0.7322 & 0.1679 & 148 & 37 & 2.4867 & 1.4531 & 1.0336 \\
OOV & 331 & 0.3988 & 0.1390 & -0.2598 & 8 & 94 & 2.6169 & 2.7188 & -0.1020 \\
action click & 845 & 0.4237 & 0.4533 & 0.0296 & 156 & 131 & 2.6067 & 2.0663 & 0.5404 \\
action input & 107 & 1.0000 & 1.0000 & 0.0000 & 0 & 0 & 2.0565 & 0.7276 & 1.3289 \\
action terminate & 40 & 1.0000 & 1.0000 & 0.0000 & 0 & 0 & 2.1804 & 0.9133 & 1.2671 \\
\bottomrule
\end{tabular}}

\par\medskip
\resizebox{0.82\linewidth}{!}{%
\begin{tabular}{llrrrr}
\toprule
Split & Model & OOV $n$ & Top-1 & NLL & MRR \\
\midrule
session & semantic CE & 331 & 0.0151 & 3.4169 & 0.1512 \\
session & semantic factorized scorer & 331 & 0.0211 & 3.2063 & 0.1711 \\
user-disjoint & user-disjoint semantic CE & 187 & 0.0000 & 3.3537 & 0.1422 \\
user-disjoint & user-disjoint semantic factorized scorer & 187 & 0.0160 & 3.2165 & 0.1584 \\
\bottomrule
\end{tabular}}

\caption{Paired A3/A4 trade-off and semantic-OOV post-hoc summaries at seed
13.  The first block reports A4 minus A3 on the same 992 rows; the second
block reports OOV metrics for the clean semantic ledgers.  Lexical buckets are
retained in the JSON artifact and are analysis-only proxies.}
\label{tab:review-diagnostics}
\end{table}

\begin{figure}[H]
\centering
\includegraphics[width=0.86\linewidth]{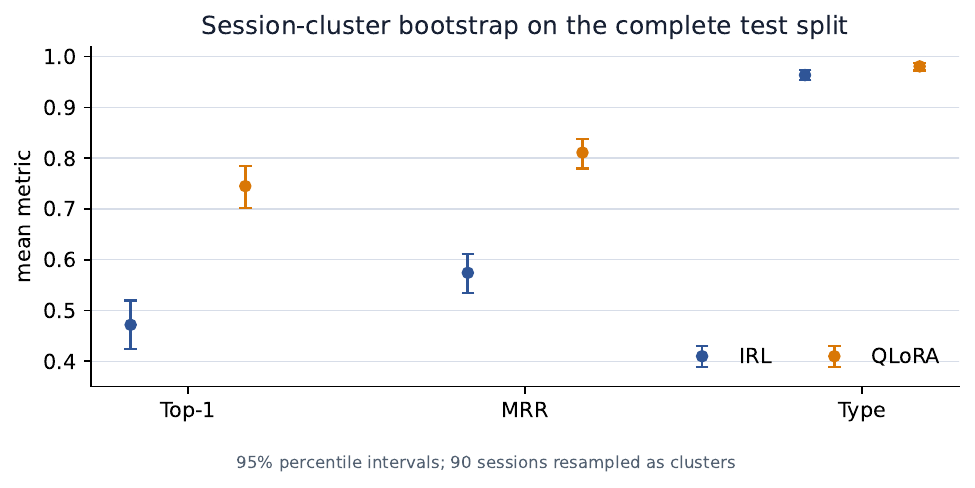}
\caption{Session-cluster bootstrap intervals computed from the complete
seed-13 held-out ledgers. Complete sessions, rather than individual rows, are
resampled; the QLoRA curve remains a directional diagnostic.}
\label{fig:session-bootstrap}
\end{figure}

To complement the seed-level standard deviations in the provisional
factorized-scorer aggregate, we also resample the complete held-out sessions in
the seed-13
ledgers.  The resulting percentile intervals are descriptive row means under
session dependence, not confidence intervals over users or training seeds.
Figure~\ref{fig:session-bootstrap} and Table~\ref{tab:session-bootstrap}
report the same 90-session bootstrap for the directional hierarchical
factorized-scorer and full-context QLoRA ledgers.

\begin{table}[H]
\centering
\small
\setlength{\tabcolsep}{4pt}
\begin{tabular}{llrrr}
\toprule
Model & Metric & Mean & 95\% cluster CI & Sessions \\
\midrule
IRL & Top-1 & 0.4718 & [0.4244, 0.5196] & 90 \\
IRL & MRR & 0.5742 & [0.5339, 0.6122] & 90 \\
IRL & Type & 0.9637 & [0.9532, 0.9737] & 90 \\
QLoRA & Top-1 & 0.7450 & [0.7019, 0.7840] & 90 \\
QLoRA & MRR & 0.8111 & [0.7797, 0.8385] & 90 \\
QLoRA & Type & 0.9808 & [0.9727, 0.9879] & 90 \\
\bottomrule
\end{tabular}

\caption{Session-cluster bootstrap intervals for seed-13 held-out metrics.
The intervals quantify descriptive variation across 90 held-out sessions and
do not replace the three-seed training variation reported in the main table.}
\label{tab:session-bootstrap}
\end{table}

\begin{table}[H]
\centering
\scriptsize
\resizebox{\linewidth}{!}{\begin{tabular}{llrccccc}
\toprule
Condition & Stratum & $n$ & Top-1 & NLL & MRR & Macro-F1 & Exact \\
\midrule
Flat IRL & in-vocab & 661 & 0.6828$\pm$0.0086 & 1.7464$\pm$0.0222 & 0.7561$\pm$0.0045 & 0.9338 & 0.6828$\pm$0.0086 \\
Flat IRL & target-OOV & 331 & 0.0735$\pm$0.0046 & 2.9519$\pm$0.0125 & 0.2263$\pm$0.0040 & 0.3267 & 0.0735$\pm$0.0046 \\
No persona IRL & in-vocab & 661 & 0.6828$\pm$0.0101 & 1.7433$\pm$0.0224 & 0.7564$\pm$0.0053 & 0.9338 & 0.6828$\pm$0.0101 \\
No persona IRL & target-OOV & 331 & 0.0745$\pm$0.0035 & 2.9532$\pm$0.0129 & 0.2263$\pm$0.0030 & 0.3267 & 0.0745$\pm$0.0035 \\
Persona IRL & in-vocab & 661 & 0.6843$\pm$0.0106 & 1.7438$\pm$0.0234 & 0.7567$\pm$0.0057 & 0.9338 & 0.6843$\pm$0.0106 \\
Persona IRL & target-OOV & 331 & 0.0725$\pm$0.0030 & 2.9523$\pm$0.0120 & 0.2260$\pm$0.0034 & 0.3267 & 0.0725$\pm$0.0030 \\
Hierarchical IRL & in-vocab & 661 & 0.6833$\pm$0.0114 & 1.7407$\pm$0.0235 & 0.7569$\pm$0.0057 & 0.9338 & 0.6833$\pm$0.0114 \\
Hierarchical IRL & target-OOV & 331 & 0.0745$\pm$0.0017 & 2.9537$\pm$0.0123 & 0.2261$\pm$0.0019 & 0.3267 & 0.0745$\pm$0.0017 \\
\bottomrule
\end{tabular}
}
\caption{Vocabulary-stratified held-out factorized-scorer results. The in-vocabulary and
target-OOV rows contain 661 and 331 test steps, respectively. The formal
2,048-token QLoRA strata are reported separately in Table~\ref{tab:appendix-sft};
the matched 384-token evaluation aggregate is in
Table~\ref{tab:appendix-sft-matched-eval384}. Bounded 1,024-row training
pilots remain excluded from paper-facing tables.}
\label{tab:appendix-strata}
\end{table}

\section{Calibration and Selective Risk}
\label{app:calibration}

\subsection{Candidate composition and calibration diagnostics}
The target-anchored evaluator fixes the target in every 16-way choice set, but
the number of same-type negatives still changes the difficulty of the
conditional ranking problem. Table~\ref{tab:appendix-composition} reports the
candidate slots aggregated by target action type. These counts describe the
evaluation contract; they are not additional training examples.

\begin{table}[H]
\centering
\small
\setlength{\tabcolsep}{4pt}
\begin{tabular}{lrrrrr}
\toprule
Target type & Rows & Click slots & Input slots & Terminate slots & Same type \\
\midrule
click & 845 & 13449 & 50 & 21 & 0.9947 \\
input & 107 & 1388 & 321 & 3 & 0.1875 \\
terminate & 40 & 600 & 0 & 40 & 0.0625 \\
\bottomrule
\end{tabular}

\caption{Candidate composition in the 992-row held-out split. ``Same type''
is the mean fraction of the 16 candidate slots whose action type matches the
target type.}
\label{tab:appendix-composition}
\end{table}

\begin{figure}[H]
  \centering
  \includegraphics[width=0.62\linewidth]{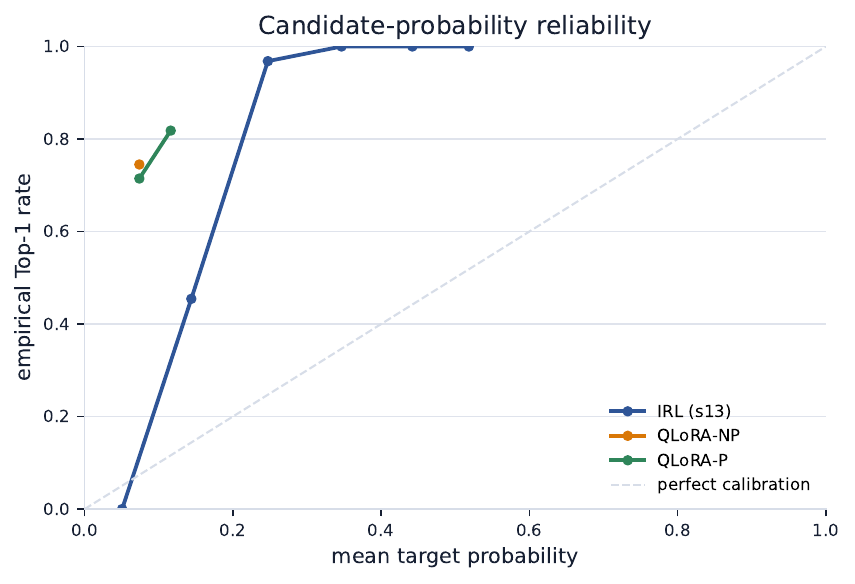}
  \caption{Reliability of the target candidate probability on the complete
  held-out split. The diagonal is perfect calibration; each curve is binned by
  predicted target probability.}
  \label{fig:appendix-calibration}
\end{figure}

Candidate probabilities are also evaluated as probabilities rather than only
as ranks. Figure~\ref{fig:appendix-calibration} shows reliability curves for
the target probability, while Table~\ref{tab:appendix-calibration} reports two
ten-bin expected calibration errors (ECEs) and the multiclass Brier score.
Target-$p$ Top-1 ECE bins the probability assigned to the gold target and
compares it with whether that target is ranked first; it is a target-anchored
diagnostic, not the usual multiclass confidence ECE. Max-confidence ECE instead
bins the largest candidate probability and compares it with whether the
argmax candidate is correct, which is the standard max-confidence multiclass
summary. These diagnostics use the same complete 992-row seed-13 records and should
not be read as evidence that the QLoRA token objective is comparable to the factorized-scorer
candidate NLL. Our multiclass Brier convention is the per-row sum over all 16
candidate classes, averaged over rows:
$\frac{1}{N}\sum_i\sum_j(p_{ij}-\mathbb{1}[j=y_i])^2$. The supplementary
JSON also retains the candidate-count-scaled mean of the same squared errors,
but the paper table reports the multiclass sum convention.

\begin{table}[H]
\centering
\small
\resizebox{\linewidth}{!}{\setlength{\tabcolsep}{5pt}
\begin{tabular}{lrrrrr}
\toprule
Condition & Target-$p$ ECE $\downarrow$ & Max-conf. ECE $\downarrow$ & Candidate NLL $\downarrow$ & Multiclass Brier $\downarrow$ & Mean target $p$ \\
\midrule
IRL (seed 13) & 0.3409 & 0.2381 & 2.1331 & 0.7721 & 0.1747 \\
QLoRA (no persona) & 0.6711 & 0.6698 & 2.6100 & 0.9156 & 0.0739 \\
QLoRA (persona) & 0.6414 & 0.6381 & 2.6235 & 0.9168 & 0.0744 \\
\bottomrule
\end{tabular}
}
\caption{Calibration diagnostics from complete 992-row seed-13 row records. Target-$p$
Top-1 ECE uses ten bins of the probability assigned to the gold target; maximum
confidence ECE uses ten bins of the largest candidate probability and argmax
correctness. Multiclass Brier is the per-row sum of the 16 squared class
errors, averaged over rows. Lower is better for all three metrics.}
\label{tab:appendix-calibration}
\end{table}

\subsection{Selective risk and failure detection}
\label{app:selective-risk}
To test whether the retained candidate probabilities support a post-hoc
abstention diagnostic, we compute a risk--coverage curve from the same
complete 992-row seed-13 ledgers. A row is correct exactly when its predicted
candidate equals the target; confidence is the maximum candidate probability.
Rows are accepted in descending confidence order, and Risk@50\% is the error
rate among the highest-confidence half of each stratum. AURC is the
trapezoidal area under the realized prefix curve, while Error AUROC scores
$1-\max_j p_{ij}$ against the binary top-1 error event.

This is a target-anchored conditional-reranking diagnostic, not an online
utility estimate, a learned abstention policy, or an open-vocabulary proposal
test: the demonstrated target remains inserted into every candidate set.
Table~\ref{tab:selective-risk} shows that the factorized scorer is selectively useful
on in-vocabulary rows but fails on target-OOV rows (AURC $=0.9342$, Risk@50\%
$=0.9096$, and Error AUROC $=0.5426$). The QLoRA ledgers retain stronger OOV
error detection in the available seed-13 comparison. The added support-aware rows
make the utility boundary explicit: the no-persona hybrid reaches Risk@50\%
$=0.0101$ overall and $0.0000$ on target-OOV rows, but its OOV stratum has no
errors so Error AUROC is undefined; the persona 384/384 hybrid has OOV
Risk@50\% $=0.1145$ and Error AUROC $=0.8047$. These are separate,
target-anchored conditional reranking diagnostics (the no-persona prior/ranker
use 2,048/1,024 tokens and the persona pair uses 384/384), not matched
comparisons or deployment-time abstention results. This is a single-seed
directional diagnostic; it does not establish open-vocabulary reliability. The
machine-readable reports and clean run receipts are retained as
the machine-readable selective-risk audit files.

\begin{table}[H]
\centering
\scriptsize
\resizebox{\linewidth}{!}{\setlength{\tabcolsep}{4pt}
\begin{tabular}{llrrrr}
\toprule
Model & Stratum & $n$ & Risk@50\% $\downarrow$ & AURC $\downarrow$ & Error AUROC $\uparrow$ \\
\midrule
IRL & Overall & 992 & 0.2359 & 0.2850 & 0.8510 \\
IRL & In-train vocab. & 661 & 0.1118 & 0.1401 & 0.8441 \\
IRL & Target-OOV & 331 & 0.9096 & 0.9342 & 0.5426 \\
QLoRA no persona & Overall & 992 & 0.0565 & 0.0857 & 0.8559 \\
QLoRA no persona & In-train vocab. & 661 & 0.0393 & 0.0537 & 0.8573 \\
QLoRA no persona & Target-OOV & 331 & 0.1687 & 0.1989 & 0.8126 \\
QLoRA persona & Overall & 992 & 0.1391 & 0.1876 & 0.7405 \\
QLoRA persona & In-train vocab. & 661 & 0.0816 & 0.1039 & 0.7653 \\
QLoRA persona & Target-OOV & 331 & 0.3193 & 0.3834 & 0.6774 \\
Support-aware no persona (ctx1024) & Overall & 992 & 0.0101 & 0.0535 & 0.8606 \\
Support-aware no persona (ctx1024) & In-train vocab. & 661 & 0.1420 & 0.1413 & 0.7595 \\
Support-aware no persona (ctx1024) & Target-OOV & 331 & 0.0000 & 0.0000 & -- \\
Support-aware persona (ctx384) & Overall & 992 & 0.3246 & 0.2929 & 0.7394 \\
Support-aware persona (ctx384) & In-train vocab. & 661 & 0.4471 & 0.4193 & 0.6863 \\
Support-aware persona (ctx384) & Target-OOV & 331 & 0.1145 & 0.1261 & 0.8047 \\
\bottomrule
\end{tabular}
}
\caption{Selective-risk and error-detection diagnostics from complete 992-row seed-13
candidate ledgers. Risk@50\% accepts the highest-confidence half; AURC is the
realized risk--coverage area; Error AUROC uses $1-\max p$ as an uncertainty
score. The support-aware rows are separate 2,048/1,024-token (no persona) and
384/384-token (persona) conditional reranking protocols; ``--'' denotes an
undefined AUROC when a stratum has one correctness class. All values are
conditional on the target anchor and directional, not evidence of open-vocabulary
or online abstention performance.}
\label{tab:selective-risk}
\end{table}

\paragraph{Reproducibility receipt.}
The review diagnostic bundle is derived without retraining.  It contains six
ordered ledgers: paired A3/A4 ledgers over 992 session-disjoint rows, two
semantic ledgers over the same 992 rows, and two semantic ledgers over the 525
user-disjoint rows.  The session-disjoint target-vocabulary split is fixed at
661 in-vocabulary versus 331 target-OOV rows; the user-disjoint split is
registered separately in its own receipt.  The machine-readable
the machine-readable review-diagnostic audit files bundle
records source/output hashes, upstream report/record/run metadata, and the
clean Git state used to materialize the summaries.

\end{document}